\documentclass[lettersize,journal]{IEEEtran}

\usepackage{amsmath,amsfonts,amssymb}
\usepackage[inline]{enumitem}
\usepackage{textcomp}
\usepackage{stfloats}
\usepackage{url}
\usepackage{verbatim}
\usepackage{graphicx}
\usepackage{multirow}
\usepackage{subfigure}
\usepackage{float}
\usepackage{comment}
\usepackage{bbm}

\usepackage{tablefootnote}

\usepackage[backend=bibtex,style=numeric,natbib=true,citestyle=numeric,sorting=none,backref=false,backrefstyle=three,maxcitenames=2]{biblatex}
\bibliography{refs}
\usepackage[breaklinks=true,colorlinks,bookmarks=false]{hyperref}
\AtBeginBibliography{\small}
\defbibfilter{appendixOnlyFilter}{
  segment=1 
  and not segment=0 
}
\usepackage[capitalise,noabbrev]{cleveref}
\usepackage[pagewise]{lineno}
\usepackage[normalem]{ulem}

\usepackage{arydshln}

\usepackage{color}
\usepackage[
monochrome
]{xcolor}

\usepackage{fancyhdr}
\fancypagestyle{copyright}{%
    \fancyhf{}
    \cfoot{\footnotesize%
        \textcopyright~2023 IEEE. Personal use of this material is permitted. Permission from IEEE must be obtained for all other uses, in any current or future media, including reprinting\slash republishing this material for advertising or promotional purposes, creating new collective works, for resale or redistribution to servers or lists, or reuse of any copyrighted component of this work in other works.    
    }
}

\begin{document}
\title{SSL4EO-S12: A Large-Scale Multi-Modal, Multi-Temporal Dataset for Self-Supervised Learning in Earth Observation}

\author{Yi Wang,~\IEEEmembership{Student Member,~IEEE}, Nassim Ait Ali Braham, Zhitong~Xiong, \IEEEmembership{Member,~IEEE}, Chenying Liu, Conrad M Albrecht,~\IEEEmembership{Member,~IEEE}, Xiao Xiang Zhu,~\IEEEmembership{Fellow,~IEEE}


\thanks{Y. Wang, N. A. A. Braham, C. Liu are with the Chair of Data Science in Earth Observation, Technical University of Munich (TUM), and the Remote Sensing Technology Institute, German Aerospace Center (DLR). Z. Xiong, X. X. Zhu are with the Chair of Data Science in Earth Observation, Technical University of Munich (TUM). C. M. Albrecht is with the Remote Sensing Technology Institute, German Aerospace Center (DLR).}

}

\markboth{ACCEPTED BY IEEE GEOSCIENCE AND REMOTE SENSING MAGAZINE, 2023}
{Shell \MakeLowercase{\textit{et al.}}: A Sample Article Using IEEEtran.cls for IEEE Journals}


\maketitle
\vspace{-2em}
\begin{abstract}
Self-supervised pre-training bears potential to generate expressive representations from large-scale Earth observation (EO) data without human annotation. However, most existing pre-training in the field is based on ImageNet or medium-size, labeled remote sensing (RS) datasets. In this paper, we share an unlabeled dataset SSL4EO-S12:
\begin{center}
\textit{
\underline{S}elf-\underline{S}upervised \underline{L}earning \underline{for} \underline{E}arth \underline{O}bservation - \underline{S}entinel-\underline{1}/\underline{2}}
\end{center}
to assemble a large-scale, global, multimodal, and multi-seasonal corpus of satellite imagery. We demonstrate SSL4EO-S12 to succeed in self-supervised pre-training for a set of representative methods: MoCo-v2, DINO, MAE and data2vec, and multiple downstream applications including scene classification, semantic segmentation and change detection. Our benchmark results prove the effectiveness of SSL4EO-S12 compared to existing datasets. The dataset, related source code, and pre-trained models are available at \url{https://github.com/zhu-xlab/SSL4EO-S12}.
\end{abstract}


\thispagestyle{copyright}

\section{Introduction}

\IEEEPARstart{S}{elf-supervised} learning (SSL) has attracted wide attention in the remote sensing (RS) community with the ability to learn generic representations from unlabeled data. Numerous studies in the literature have proven the potential of SSL in Earth observation (EO) beyond natural images \cite{wang2022self1}. Despite the focus SSL for EO receives, only limited effort is dedicated to providing large-scale datasets and benchmarks for pre-training. On the one hand, relying on computer vision datasets like ImageNet \citep{deng2009imagenet} is not a preferred option due to the domain gap. On the other hand, while RS datasets like SEN12MS \cite{Schmitt2019} or SeCo \cite{manas2021seasonal} exist, they are limited by geospatial overlap, sparse geographical distribution, or lack diversity in seasonal or multimodal information. Therefore, big EO-specific datasets for unsupervised pre-training are necessary to be developed.

In this work, we introduce a large-scale, globally distributed, multi-temporal and multi-sensor dataset \textit{SSL4EO-S12: Self-Supervised Learning for Earth Observation - Sentinel-1/2}. The dataset samples 250K locations around the globe, each providing Sentinel-2 L1C, Sentinel-2 L2A, and Sentinel-1 GRD images with four snapshots from different seasons (in total 3 million 2640m$\times$2640m patches). Additionally, we guarantee optimal geospatial coverage by avoiding the overlap of the randomly sampled locations. This renders SSL4EO-S12 the largest and most generic multi-spectral/SAR dataset in the RS literature \cite{xiong2022earthnets}.

We demonstrate the potential of SSL4EO-S12 dataset through a series of extensive experiments. Specifically, we evaluate four representative SSL algorithms---namely: MoCo \cite{he2020momentum}, DINO \cite{caron2021emerging}, MAE \cite{he2021masked}, and data2vec \cite{baevski2022data2vec}---on three different downstream tasks: scene classification, semantic segmentation and change detection. Our results indicate that pre-training on SSL4EO-S12 improves the downstream performance compared to existing datasets. Moreover, our ablation studies prove the benefits of RS-specific data augmentations including multi-sensor, multi-temporal and atmospheric correction.

\begin{figure*}
  \centering
  \includegraphics[width=1.0\linewidth]{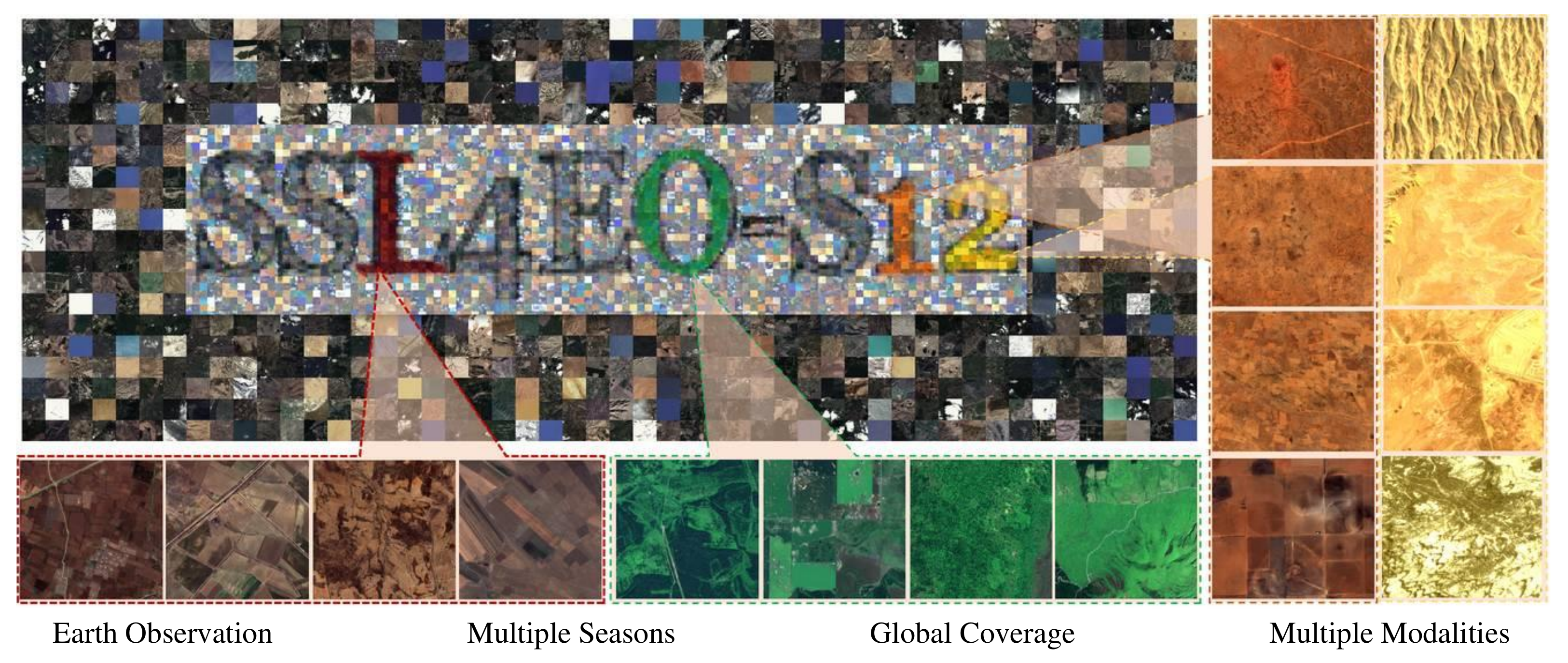}
  \caption{Sample images of SSL4EO-S12 dataset assembled.}
  \label{fig:ssl4eo-s12-logo}
\vspace{-1em}
\end{figure*}


\section{Related work}
\noindent{\textbf{Self-supervised learning}} \hspace{0.3em}
Over the past years, self-supervised learning (SSL) has reached important milestones in computer vision, especially through contrastive methods with joint-embedding architectures. These methods get trained to promote similarity between augmented views of the same input, thereby enforcing invariance to data augmentation. Several families of such methods emerge:
1) contrasting negative samples for which the representations are encouraged to be dissimilar \cite{he2020momentum}; 2) Knowledge distillation between an asymmetric teacher-student network \cite{caron2021emerging}; 3) redundancy reduction among the embedding dimensions; 4) clustering latent features to common prototypes from different views \cite{caron2020unsupervised}. Meanwhile, recent developments in masked image modeling (MIM) reveal promising results in generative methods, which reconstruct the masked input at pixel-\cite{he2021masked} or feature-\cite{baevski2022data2vec} level. 

We benchmark four representative methods MoCo \cite{he2020momentum}, DINO \cite{caron2021emerging}, MAE \cite{he2021masked}, and data2vec \cite{baevski2022data2vec} on the proposed dataset. This way, we cover a reasonably diverse set of representative methods from different categories: MoCo contrasts negative samples, DINO represents a distillation method, MAE is based on masked reconstruction, and data2Vec combines the masking mechanism with a joint-embedding architecture. 

\vspace{0.5em}
\noindent{\textbf{Pre-training datasets}} \hspace{0.3em}
Pre-trained models on ImageNet are widely used for various computer vision tasks. However, this is less appropriate in the context of RS:
1) RS images are not object-centric; 2) there exist various types of sensors in RS; 3) temporal effects yield variations on the ground surface. Therefore, EO-specific datasets are needed to provide the above in-domain knowledge. The literature has proven the benefits of pre-training on existing labeled RS datasets \cite{neumann2020training,wang2022self}, yet there are limitations such as class bias, and temporal and geographical coverage.

Consequently, there is a need for large-scale pre-training datasets in RS. Two datasets closely related to our efforts are SEN12MS \cite{Schmitt2019} and SeCo \cite{manas2021seasonal}. However, SEN12MS is limited by temporal coverage, SeCo has only optical data, and both datasets contain strongly overlapping patches which limit the geospatial coverage. With the above in mind, our proposed SSL4EO-S12 dataset provides an improved spatio-temporal coverage by sampling more locations and removing overlapping patches, enclosing multiple seasons, and including Sentinel-1 as well as two Sentinel-2 products (Table \ref{tab:datasets}).     

\begin{table*}[]
\centering
\caption{Summary of popular medium-resolution pre-training datasets in remote sensing. LC: land cover.}
\label{tab:datasets}
\scalebox{1.0}{
\begin{tabular}{lccccccc}
\hline
dataset   & spatial cover   & temporal cover     & modality             & overlap     & patch size  & $\#$ of locations & $\#$ of patches       \\ \hline \hline
BigEarthNet \cite{sumbul2019bigearthnet} 
& Europe          & 1 timestamp           & SAR/optical & no & 120$\times$120 & 600K & 1.2M      \\
SEN12MS \cite{Schmitt2019}               
& global & 1 timestamp           & SAR/optical/LC & yes         & 256$\times$256 & 180K & 540K       \\
SeCo \cite{manas2021seasonal}            
& global & 5 timestamps & optical              & yes         & 264$\times$264 & 200K & 1M       \\
SSL4EO-S12  & global & 4 timestamps & SAR/optical*2 & $\sim$no & 264$\times$264 & 250K & 3M \\ \hline
\end{tabular}
}
\vspace{-1em}
\end{table*}

\section{SSL4EO-S12 Dataset}
\label{sec:ssl4eo-s12}

\subsection{Data curation \& assembly}
\label{sec:datacuration}
The SSL4EO-S12 dataset (\cref{fig:ssl4eo-s12-logo}) exploits openly available SAR/optical satellite data collected by the European Space Agency's Sentinel mission. Following a well-organized baseline provided by SeCo \cite{manas2021seasonal}, we utilize the Google Earth Engine \citep{gorelick2017google} to download and process the data. We filter image patches to retrieve from the 10,000 most populated cities\footnote{\url{https://simplemaps.com/data/world-cities}} in the world (top-10k) to guarantee reasonable global coverage. To obtain diverse land cover, we sample 251,079 locations close by the cities following a Gaussian distribution peaking at the city center and standard deviation of 50km---assuming most of the variability cast to the downtown and suburbs of cities \citep{manas2021seasonal}. At each location, we download 4 images drawn from four annual seasons to capture seasonal variation. We search for Sentinel-2 tiles with a cloud coverage lower than 10\%. We also filter out most overlapping patches with an efficient grid search strategy. In total, we obtain about one million S1-GRD/S2-L1C/S2-L2A image triplets.

\textbf{Data identification.} The collection of SSL4EO-S12 differs from SeCo mainly by introducing overlap filtering and multiple sensors (\textbf{bold} below). The workflow is shown as follows:

\begin{enumerate}
\small
    \item Uniformly sample one city from top-10k populated cities;
    \vspace{0.2em}
    \item Sample one location from a Gaussian distribution with a standard deviation of 50km around the city center;
    \vspace{0.2em}
    \item \textbf{Check if a 2640m$\times$2640m image patch centered around that location has significant overlap with previous patches. If not, continue to 4, otherwise return to 1;}
    \vspace{0.2em}
    \item For a 30-day interval around four reference dates (Mar 20, Jun 21, Sep 22, Dec 21) in 2021 (additionally look for 2020 as a buffer), check if there exist Sentinel-2 tiles with less than 10\% of cloud coverage (\textbf{for both L1C and L2A}) and corresponding \textbf{Sentinel-1 GRD tiles};
    \vspace{0.2em}
    \item If there exist valid Sentinel-1/2 tiles close to all the four dates, process and download them into curated image patches, otherwise return to 1.
\end{enumerate}

\textbf{Overlap filtering.} A simple way to check significant overlap between two patches is to calculate the distance between the two centers. If the distance is smaller than 3/4 the width of a patch, there is a non-negligible overlap (>25\%). Naively, we need to execute this computation for every new patch relative to all existing patches. However, this becomes inefficient when the number of patches grows large, 250k+ for us. 
Therefore, we employ a grid search strategy to perform efficient overlap filtering. Instead of calculating
the distance to all previous patches, we distribute the patch center coordinates into 360x180 geographical longitude-latitude, one-by-one-degree grids. For each new patch, we convert the center coordinates into integer grid coordinates. Subsequently, we search for existing patches within this grid cell
and exclusively calculate distances to those local patches. Assuming potential overlap of sampled patches
from distinct grid cells is statistically negligible, we significantly reduce computing time compared
to a global overlap search. Indeed, for SSL4EO-S12 we record an overlap for approx. 3\% tiles of densely populated Tokyo, 1.5\% in Chicago, and below 1\% for locations such as Bejing, Munich, Kampala, and Brasilia.

\begin{figure}
  \centering
  \includegraphics[width=0.9\linewidth]{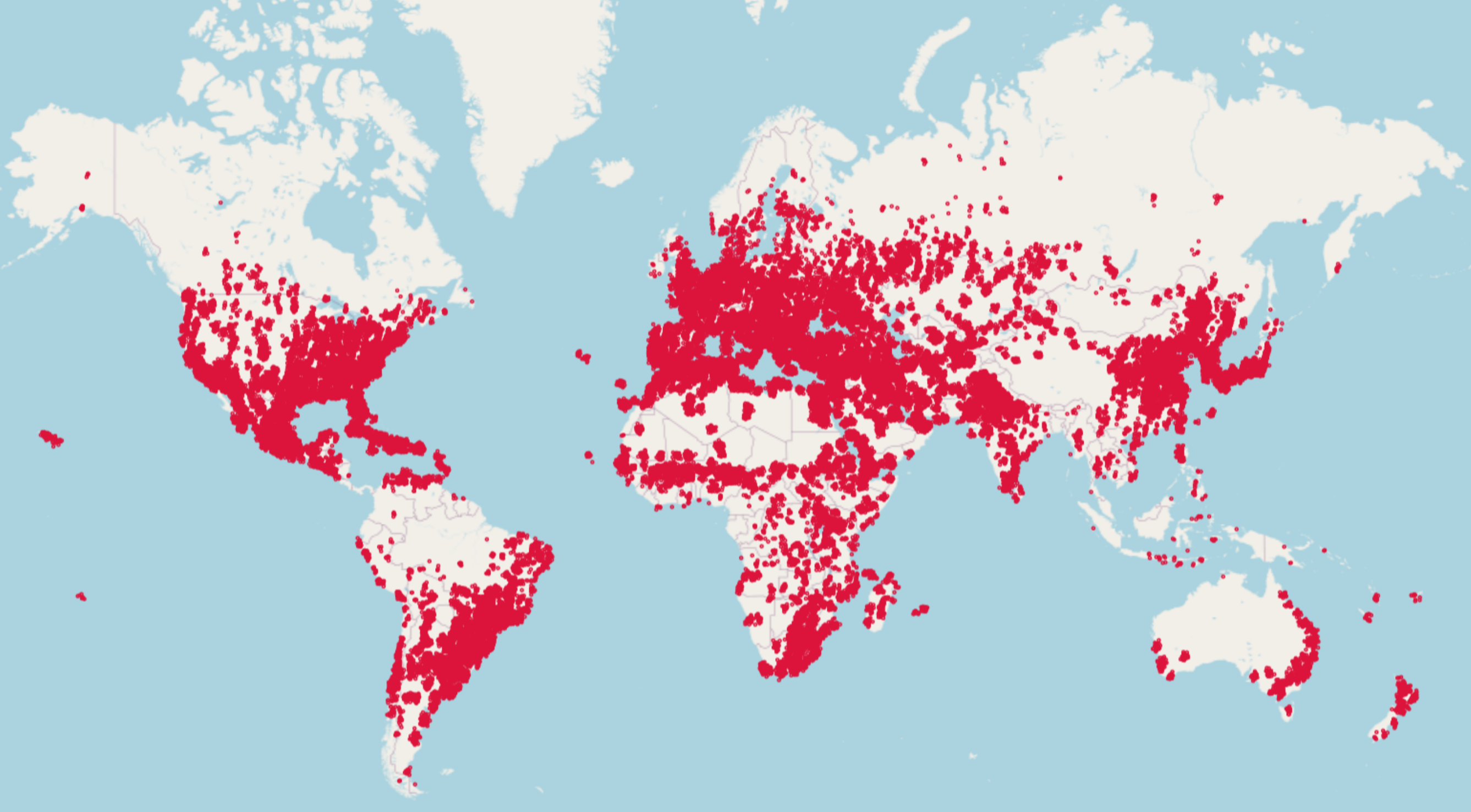}
  \caption{Geographical distribution of SSL4EO-S12 dataset.}
  \label{fig:globe}
\end{figure}

\begin{figure}[t!]
  \centering
  \includegraphics[width=0.45\linewidth]{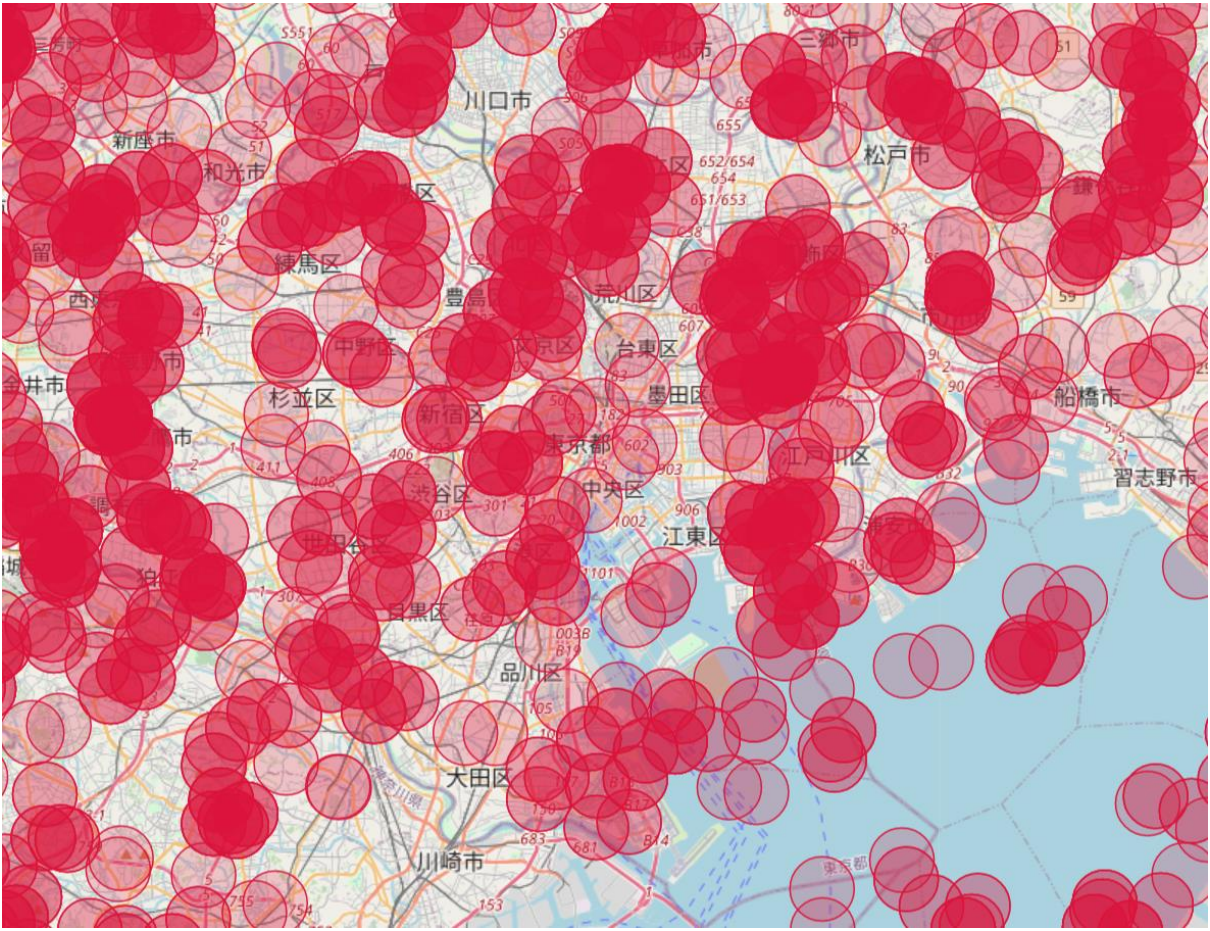}
  \includegraphics[width=0.45\linewidth]{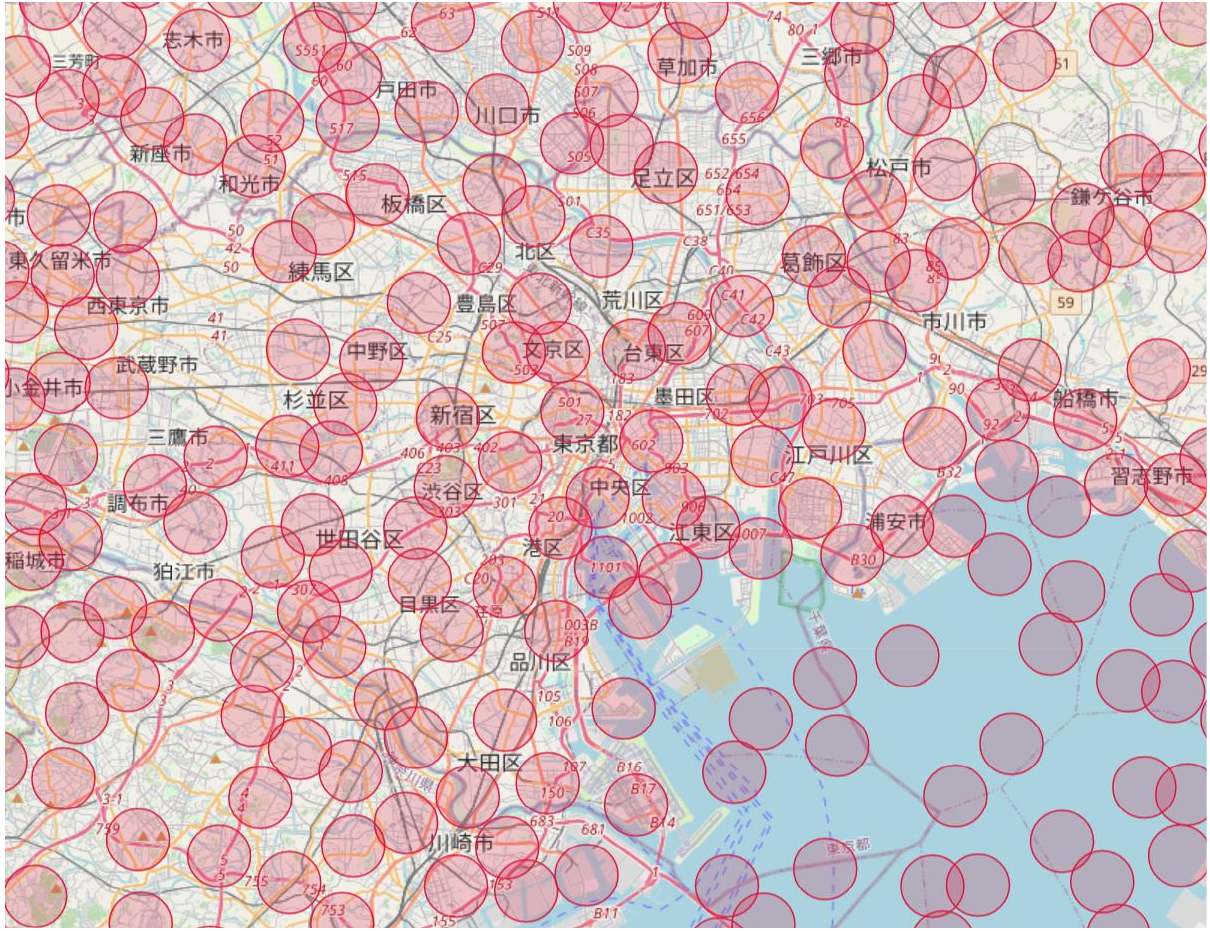}  
  \caption{Image patches without (left) and with (right) overlap filtering in Tokyo metropolitan area.
  We plot red circles of radius 1.32km (132 pixels) for better visualization.}
  \label{fig:overlap}
  \vspace{-1em}
\end{figure}

\vspace{-1em}
\subsection{Data characteristics \& volume}

The presented SSL4EO-S12 dataset contains 251,079 globally distributed Sentinel-1 dual-pol SAR, Sentinel-2 top-of-atmosphere multispectral, and Sentinel-2 surface reflectance multispectral triplets over four seasonal timestamps. As of summer 2022, SSL4EO-S12 constitutes the biggest geospatial-temporal, multimodal dataset in terms of medium-resolution PolSAR and multi-spectral imagery serving more than 3 million images. The total data volume equates to an uncompressed size of 
$251,079\times4\times\left[2\cdot4B+(13+12)\cdot2B\right]\times264^2\approx3.7TB~.$

\Cref{fig:globe} depicts the geospatial distribution of the SSL4EO-S12 dataset, highlighting the dense coverage across the globe. \Cref{fig:overlap} depicts the effect of overlap filtering around Tokyo area. 

\vspace{-0.5em}
\section{Experimental setup}
\label{sec:setup}
We evaluate SSL4EO-S12 dataset by self-supervised pre-training and transfer learning on RS downstream tasks. Specific implementation details are provided in the appendix.

\subsection{Self-supervised pre-training} We perform pre-training using four representative SSL methods: \textit{MoCo-v2/v3} \citep{chen2020improved,chen2021empirical}, \textit{DINO} \citep{caron2021emerging}, \textit{MAE} \citep{he2021masked}, and \textit{data2vec} \citep{baevski2022data2vec}.
We pre-train ResNet \citep{he2016deep} backbones with MoCo(-v2) and DINO, and Vision Transformer (ViT) \citep{dosovitskiy2020image} backbones for all four SSL methods listed above. Unless explicitly noted, Sentinel-2 L1C is used for pre-training.
To utilize multi-temporal information, we use \texttt{RandomSeasonContrast} as a data augmentation strategy, i.e., for MoCo and DINO, the input views are randomly picked from two seasons. For MAE and data2vec, one random season is assigned for each patch.

Pre-training one ResNet/ViT model for 100 epochs takes 7--25 hours on 4 NVIDIA A100 GPUs, as shown in Table \ref{tab:pretraintime}.

\begin{table}[h!]
\centering
\caption{100 epoch training time of the studied SSL methods.}
\label{tab:pretraintime}
\begin{tabular}{ccccc}
\hline
& MoCo & DINO & MAE & data2vec            \\ \hline \hline
ResNet50 & 18h  & 25h  & -   & -                   \\
ViT-S/16 & 24h  & 25h  & 7h  & 14h \\ \hline
\end{tabular}
\vspace{-1em}
\end{table}

\subsection{Transfer learning} The pre-trained models are transferred to various downstream tasks. For
\begin{itemize}
    \item \textit{scene classification}, we evaluate EuroSAT \citep{helber2019eurosat} (single-label land cover classification), BigEarthNet \citep{sumbul2019bigearthnet} (multi-label land cover classification), and So2Sat-LCZ42 \citep{zhu2020so2sat} (local climate zone classification, culture-10 version).
    \item \textit{semantic segmentation}, we include DFC2020 \citep{rha7-m332-19} (land cover segmentation) and OSCD \citep{daudt2018urban} (change detection).
\end{itemize}
We perform commonly used linear probing (freezing the pre-trained encoder) and fine-tuning for the downstream tasks. The results are reported in percentage scores.

\section{Benchmark results}
\label{sec:benchmark}

\begin{table*}[t!]
\centering
\caption{Linear probing results for EuroSAT, BigEarthNet (BE) and So2Sat-LCZ42 (10\% and 100\% labels). We report overall accuracy for EuroSAT and So2Sat-LCZ42, and mean average precision (micro) for BigEarthNet. Two backbone networks get trained: ResNet-50 (RN50) and a small(-S) embedding dimension Vision Transformer(ViT) subdividing input patches into 16x16 tiles(/16). \textbf{Bold} values indicate best per-column performance.}
\label{tab:benchmark-lc}
\scalebox{1.0}{
\begin{tabular}{|c||cc|cccc|cccc|}
\hline
\multicolumn{1}{|c}{\footnotesize downstream dataset} & \multicolumn{2}{c}{EuroSAT}   & \multicolumn{2}{c}{BE-10\%}   & \multicolumn{2}{c}{BE-100\%}  & \multicolumn{2}{c}{So2Sat-10\%} & \multicolumn{2}{c|}{So2Sat-100\%} \\
\footnotesize model~\textbackslash~backbone    & RN50          & ViT-S/16      & RN50          & ViT-S/16      & RN50          & ViT-S/16      & RN50       & ViT-S/16       & RN50            & ViT-S/16       \\ \hline
\texttt{rand.init.}            & 82.0          & 81.3          & 61.0          & 60.0          & 60.0          & 60.0          & 48.8           & 49.3           & 49.0            & 50.2           \\
\textit{supervised}               & \textbf{98.0} & 96.7          & \textbf{83.4} & 81.3          & \textbf{88.7} & \textbf{87.4} & 57.5           & 59.7           & 57.5            & 59.3           \\ \cdashline{1-11}
MoCo              & \textbf{98.0} & \textbf{97.7} & 82.1          & \textbf{82.3} & 84.2          & 83.1          & \textbf{61.3}  & 59.6           & \textbf{61.8}   & 62.2           \\
DINO              & 97.2          & \textbf{97.7} & 82.0          & 81.7          & 83.9          & 83.4          & 55.5           & \textbf{60.9}  & 57.0            & \textbf{62.5}  \\ \cdashline{1-11}[.4pt/3pt]
MAE               & -             & 94.1          & -             & 77.5          & -             & 78.2          & -              & 59.5           & -               & 60.0           \\
data2vec          & -             & 96.9          & -             & 77.3          & -             & 79.4          & -              & 58.2           & -               & 59.7           \\  \hline

\end{tabular}
}

\vspace{1ex}

\caption{Fine-tuning results for EuroSAT, BigEarthNet, and So2Sat-LCZ42. All beat \textit{supervised} training, cf.\ \cref{tab:benchmark-lc}.}
\label{tab:benchmark-ft}
\scalebox{1.0}{
\begin{tabular}{|c||cc|cccc|cccc|}
\hline
\multicolumn{1}{|c}{\footnotesize downstream dataset} & \multicolumn{2}{c}{EuroSAT}   & \multicolumn{2}{c}{BE-10\%}   & \multicolumn{2}{c}{BE-100\%}  & \multicolumn{2}{c}{So2Sat-10\%} & \multicolumn{2}{c|}{So2Sat-100\%} \\
\footnotesize model~\textbackslash~backbone   & RN50          & ViT-S/16      & RN50          & ViT-S/16      & RN50          & ViT-S/16      & RN50           & ViT-S/16       & RN50            & ViT-S/16       \\ \hline
MoCo              & \textbf{99.1} & 98.6          & 86.2          & 86.1          & \textbf{91.8} & 89.9          & 60.4           & 61.2           & 60.9            & 61.6           \\
DINO              & \textbf{99.1} & 99.0          & \textbf{87.1} & \textbf{86.9} & 90.7          & \textbf{90.5} & \textbf{63.2}  & 61.5           & \textbf{63.6}   & 62.2           \\ \cdashline{1-11}[.4pt/3pt]
MAE               & -             & 98.7          & -             & 84.8          & -             & 88.9          & -              & 60.8           & -               & 63.9           \\
data2vec          & -             & \textbf{99.1} & -             & 85.6          & -             & 90.3          & -              & \textbf{63.2}  & -               & \textbf{64.8}  \\ \hline
\end{tabular}
}
\vspace{-1em}
\end{table*}

\subsection{Classification}

\subsubsection{Comparison of SSL methods} We first benchmark different SSL methods through linear probing on EuroSAT, BigEarthNet, and So2Sat-LCZ42. As detailed in \cref{tab:benchmark-lc}, all methods outperform random initialization (\texttt{rand.init.}) by a substantial margin.
As expected, linear probing on BigEarthNet with all labels performs worse than fully \textit{supervised} training. Promisingly, the gap stays below 5\%. On small datasets like BigEarthNet with 10\% labels or EuroSAT, linear probing provides results comparable to supervised training within approx.\ $\pm1\%$. 
The trends are slightly different for So2Sat-LCZ42, where the training and testing sets are built upon different cities with a challenging geographical split. Because of this significant domain shift, adding labeled training data does not necessarily improve the testing performance. In fact, fitting the training data distribution does not guarantee out-of-distribution generalization. Nevertheless, the best pre-trained models with linear probing beat the supervised baseline by at least 1\% up to about 4\%.

Furthermore, we benchmark fine-tuning results in \cref{tab:benchmark-ft}. All self-supervised methods outperform supervised learning with a margin from 1\% to 6\%. Top SSL-models score 99.1\% on EuroSAT (MoCo/DINO) and over 90\% on BigEarthNet (MoCo/DINO). Comparing linear probing and fine-tuning results, one interesting phenomenon shows up: in linear probing contrastive methods (MoCo and DINO) consistently score better than their image-masking (MAE and data2vec) counterparts.

\subsubsection{Comparison of pre-training datasets} To compare SSL4EO-S12 with other RS pre-training datasets, we report corresponding linear probing results pre-trained with MoCo-v2 (ResNet50 backbone) in \cref{tab:dataset-comp-lc}. Similar to SSL4EO-S12, \texttt{RandomSeasonContrast} is used to pick one timestamp image for each geospatial patch in SeCo dataset. In the first set of comparison, we use RGB bands only. SSL4EO-S12 significantly outperforms ImageNet by about 10\%, SeCo by about 6\%, and SEN12MS by 1.7\% to 3.5\%.

\begin{table}[h]
\centering
\caption{Comparison of different pre-training datasets under linear probing evaluation. \textit{italic} means cited from the literature.}\label{tab:dataset-comp-lc}
\scalebox{1.0}{
\begin{tabular}{lccccc}
\hline
dataset   & EuroSAT & BE-10\% & BE-100\% \\ \hline\hline
\it ImageNet (RGB) \cite{manas2021seasonal} & \it 86.4 & \it 70.5   &\it  71.8  \\
\it SeCo (RGB) \cite{manas2021seasonal}    &\it  89.5 & \it 74.5    &\it  76.3  \\
SEN12MS (RGB)       & 94.9        & 76.6            & 79.6  \\
SSL4EO-S12 (RGB)    & \textbf{96.6}     & \textbf{80.1}      & \textbf{82.3}  \\ \hdashline
SeCo\tablefootnote{The available SeCo data at \url{https://github.com/ServiceNow/seasonal-contrast} has only about 160k geographical patches (instead of 200k in the paper), which may affect our reproduced performance.} (all bands)   & 89.2   & 73.7    & 76.6 \\
SEN12MS (all bands)   & 95.5   & 79.6    & 82.1   \\
BigEarthNet (all bands)   & 94.4   & 80.6    & 83.9  \\
SSL4EO-S12 (all bands) & \textbf{98.0}   & \textbf{82.1}    & \textbf{84.2}  \\ \hline
\end{tabular}
}
\end{table} 

In a second set of experiments we evaluate all multispectral bands.
Results indicate consistent performance gain as in RGB setting comparing SSL4EO-S12 with SEN12MS and SeCo. In addition, pre-training on SSL4EO-S12 outperforms BigEarthNet on itself and EuroSAT (both are EU only). This proves SSL4EO-S12's benefits to improve model transferability by learning valuable knowledge from a larger scale and wider geographical coverage.

\subsubsection{Comparison of different amounts of labels} 
\begin{figure*}[]
\centering
\begin{minipage}[c]{.53\linewidth}
 \includegraphics[width=\linewidth]{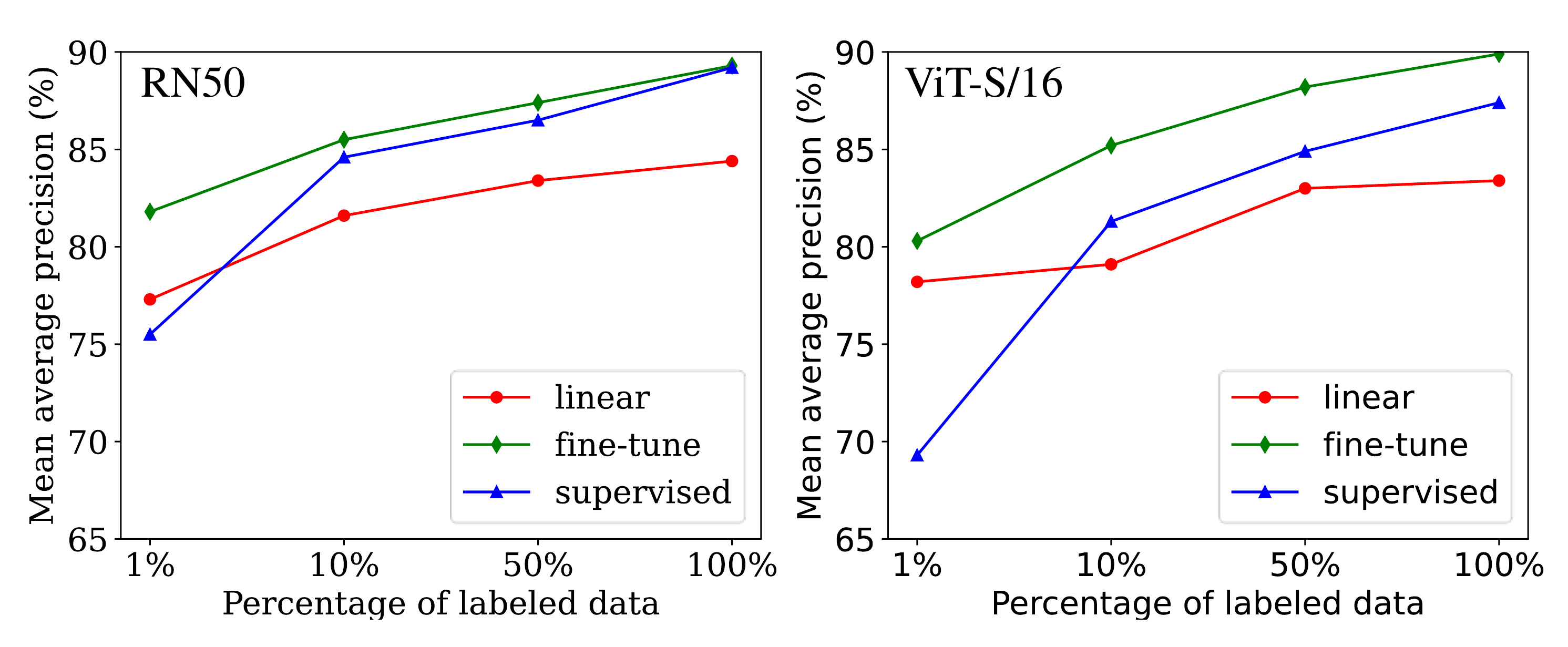}
\end{minipage}
\scalebox{0.8}{
\begin{tabular}{cccccc}
\hline
   \multicolumn{2}{c}{BE label percentage}            & 1\%        & 10\%       & 50\%       & 100\%                     \\ \hline \hline
\multirow{3}{*}{\rotatebox{90}{RN50}}     & Linear     & 75.9          & 82.1          & 82.7          & 84.2                         \\
                          & Fine tune  & \textbf{80.3} & \textbf{86.2} & \textbf{87.7} & \textbf{91.8}                \\
                          & \textit{supervised} & 75.7          & 83.4          & 85.2          & 88.7                         \\ \hline
\multirow{3}{*}{\rotatebox{90}{\small ViT-S/16}} & Linear     & 78.2          & 82.3          & 83.0          & 83.1 \\
                          & Fine tune  & \textbf{78.9}            & \textbf{86.1} & \textbf{88.2} & \textbf{89.9}                \\
                          & \textit{supervised} & 69.3          & 81.3          & 84.9          & 87.4                         \\ \cline{1-6} 
\end{tabular}
}
\caption{BigEarthNet (BE) performance depending on amount of labels available to train downstream task. We report linear probing and fine-tuning results with ResNet50 and ViT-S/16 encoders pre-trained using MoCo-v2.}
\label{fig:lowshot}
\vspace{-1em}
\end{figure*}

\Cref{fig:lowshot} visualizes performance results of transfer learning on BigEarthNet with a varying fraction of labeled samples. Compared to the supervised baseline, self-supervised pre-training on SSL4EO-S12 provides significant benefits when the amount of labeled samples is limited. In fact, fine-tuning on 10\% of the labels outperforms 50\%-labels supervised training; and with ViT-S/16, fine-tuning on 50\% of the labels outperforms 100\%-labels supervised training.
\vspace{-0.5em}

\subsection{Segmentation}
\label{sec:segmentation}

\subsubsection{Land cover segmentation} We use DFC2020 \citep{rha7-m332-19} dataset to evaluate land cover semantic segmentation. We pre-train ResNet50 with MoCo-v2 on SSL4EO-S12 L1C products, and fine-tune a DeepLabv3+ \citep{chen2018encoder} for segmentation. \Cref{tab:dfc2020} lists results with notable improvements when compared to SeCo pre-training. However, SSL4EO-S12 performs worse than SEN12MS in average accuracy (AA) and mean intersection over union (mIoU). This can be expected, since DFC2020 was built with direct reference to SEN12MS and they have similar data distribution. Nevertheless, the results are still comparable, proving again the transferability of the proposed dataset.
\vspace{-0.5em}
\begin{table}[h]
\centering
\caption{DFC2020 land cover segmentation results.}
\label{tab:dfc2020}
\begin{tabular}{cccc}
\hline
dataset         & OA & AA & mIoU \\ \hline \hline
\texttt{rand.init.}   & 81.97 & 56.46 & 42.11 \\
SeCo   & 87.31 & 57.05 & 49.68 \\
SEN12MS & 88.64 & \textbf{67.69} & \textbf{54.83}  \\
SSL4EO-S12 & \textbf{89.58} & 64.01 & 54.68  \\ 
\hline
\end{tabular}
\end{table}

\subsubsection{Change detection} We evaluate the pre-trained models for change detection on the OSCD \citep{daudt2018urban} dataset. We pre-train ResNet50 with MoCo-v2 on SSL4EO-S12 L1C products, freeze the backbone, and fine-tune a U-Net \citep{ronneberger2015u} for segmentation. The differences in feature maps between two timestamps are input to the network. As \cref{tab:oscd} indicates, pre-training on SSL4EO-S12 yields superior performance in recall and F1-score when referenced to SeCo and SEN12MS. While SSL4EO-S12 performs worse in precision, this is due to the significant class unbalance that predicting all pixels as unchanged would result in a good precision score.
\vspace{-0.5em}
\begin{table}[t]
\centering
\caption{OSCD change detection results.}
\label{tab:oscd}
\begin{tabular}{cccc}
\hline
dataset             & precision & recall & F1    \\ \hline \hline
\texttt{rand.init.}       & 72.31     & 13.75  & 23.10 \\
SeCo  & \textbf{74.85} & 17.47   & 28.33 \\
SEN12MS   & 74.67  & 19.26  & 30.62 \\
SSL4EO-S12     & 70.23     & \textbf{23.38}  & \textbf{35.08} \\
\hline
\end{tabular}
\vspace{-1em}
\end{table}

\section{Additional studies}
\label{sec:add-study}

We complete our benchmark by reporting a set of additional results to document key characteristics of the SSL4EO-S12 dataset, namely: multi-temporal, multimodal, multi-product-level, and data scale. For all studies, we pre-train ResNet50 with MoCo-v2 as a common setting.
\vspace{-0.5em}
\subsection{Ablation studies}

\subsubsection{Benefits of multimodality} While \cref{sec:benchmark} employs only optical data for fair comparison to existing literature, we highlight the benefits of multimodal pre-training in this section. We integrate SAR data by early fusion, and use \texttt{RandomSensorDrop} \citep{wang2022self} as an additional data augmentation strategy. During training, the model gets fed random combinations of SAR/optical patches, thus learning both inner- and inter-modality representations. Then, the pre-trained model gets transferred to different scenarios where either both modalities or a single one is available. We compare multimodal pre-training (MM) to uni-modal pre-training (S1/2) on BigEarthNet. \Cref{tab:modality} presents results with notable improvement of 1\%--3\% for 100\% and 1\% label splits. While single-modality pre-training already works well for both Sentinel-2 and Sentinel-1 data, pre-training exploiting both modalities further improves performance.
\begin{table}[h!]
\caption{Linear probing results of multimodal SSL. MoCo-S1/2 represents pre-training with one single modality, and MoCo-MM represents pre-training with both modalities.}
\label{tab:modality}
\centering
\begin{tabular}{ccccccc}
\hline
downstr./  & \multicolumn{3}{c}{BE-1\%}                  & \multicolumn{3}{c}{BE-100\%}                    \\
model      & S1            & S2            & S1+S2          & S1            & S2            & S1+S2          \\ \hline \hline
MoCo-S1/2  & 71.1          & 75.9          & --             & 75.9          & 84.2          & --            \\
MoCo-MM    & \textbf{73.3} & \textbf{76.7} & \textbf{76.8}  & \textbf{79.5} & 85.1          & 85.2          \\
supervised & 66.7          & 75.7          & 76.4           & 77.2          & \textbf{88.7} & \textbf{88.9}  \\ \hline
\end{tabular}
\end{table}

\subsubsection{Ablation of seasonal information} We evaluate the effectiveness of multi-temporal information by replacing seasonal \textit{augment}ation (cf.\ \cref{sec:setup}) by \textit{random} season: the same randomly selected season for the two positive views; and \textit{fixed} season: the same season for each patch during training. We pre-train on a 50k subset of SSL4EO-S12, and evaluate on BigEarthNet-10\% and EuroSAT. \cref{tab:season} clearly proves the benefits of seasonal augmentation.
\vspace{-0.5em}
\begin{table}[h!]
\centering
\caption{Linear probing results of multi-temporal ablation study.}
\label{tab:season}
\begin{tabular}{ccc}
\hline
downstr./season & BE-10\%    & EuroSAT    \\ \hline \hline
fixed           & 75.1       & 93.1       \\
random          & 76.7 & 94.0 \\
augment         & \textbf{77.6} & \textbf{96.2} \\ \hline
\end{tabular}
\vspace{-1em}
\end{table}

\subsubsection{Atmospheric correction as data augmentation} The motivation to include Sentinel-2 L1C and L2A products in SSL4EO-S12 is to match corresponding downstream tasks. However, these product levels with or without atmospheric correction can also be considered natural data augmentation for SSL. Accordingly, we conduct an ablation study on a 50k SSL4EO-S12 subset utilizing Sentinel-2 L1C, L2A or both (L1C+L2A). \Cref{tab:level} summarizes our findings:
1) models pre-trained on the same product level as the downstream task have a slight edge ($\sim1\%$) over models trained on the other product level, and 2) pre-training on both modalities generates a notable improvement of up to 4\% compared to pre-training on single modality.
\vspace{-0.5em}
\begin{table}[h!]
\caption{Linear probing results of different product levels of Sentinel-2.}
\label{tab:level}
\centering
\begin{tabular}{ccc}
\hline
product & BE-10\% (L2A)                        & EuroSAT (L1C)                        \\ \hline \hline
L1C     & 74.0                                 & 93.1                          \\
L2A     & 75.1                          & 92.0                                 \\
L1C+L2A & \textbf{78.0} & \textbf{93.8} \\ \hline
\end{tabular}
\end{table}

\subsubsection{Impact of pre-training scale} An aspect relevant to large-scale data mining in Earth observation is scaling of results with training data volume: why don't we add more images to SSL4EO-S12? One reason concerns computational costs. We believe the current dataset (1M patches for each Sentinel product) is comparable to the scale of ImageNet, and can serve as a good baseline in remote sensing for further development. Moreover, as observed by \cite{cole2022does}, saturating downstream performance kicks in beyond 500k pre-training images on ImageNet, with 250k images yielding acceptable results with as little as 1-2\% accuracy loss. We observe such a trend in our dataset, too. As demonstrated by \cref{tab:dataset-size}, we pre-train on various amounts of data to report linear probing results for BigEarthNet-10\%. While 50\% (500K) or less pre-training data yields significant performance drops, there's little diminishing gaps from 75\% (750K) on. Note this saturation effect depends also on the model size.
\vspace{-0.5em}
\begin{table}[h!]
\caption{Linear probing results on BigEarthNet-10\% for various Sentinel-2 L1C pre-training data sizes.}
\label{tab:dataset-size}
\centering
\begin{tabular}{lccccc}
\hline
data size     & 100K & 250K & 500K & 750K & 1M \\ \hline \hline
accuracy (\%) & 64   & 73   & 78   & 81   & 82 \\ \hline
\end{tabular}
\end{table}

\vspace{-1.5em}
\subsection{Representation visualization}
\begin{figure}[h]
    \centering
    \includegraphics[width=0.45\linewidth]{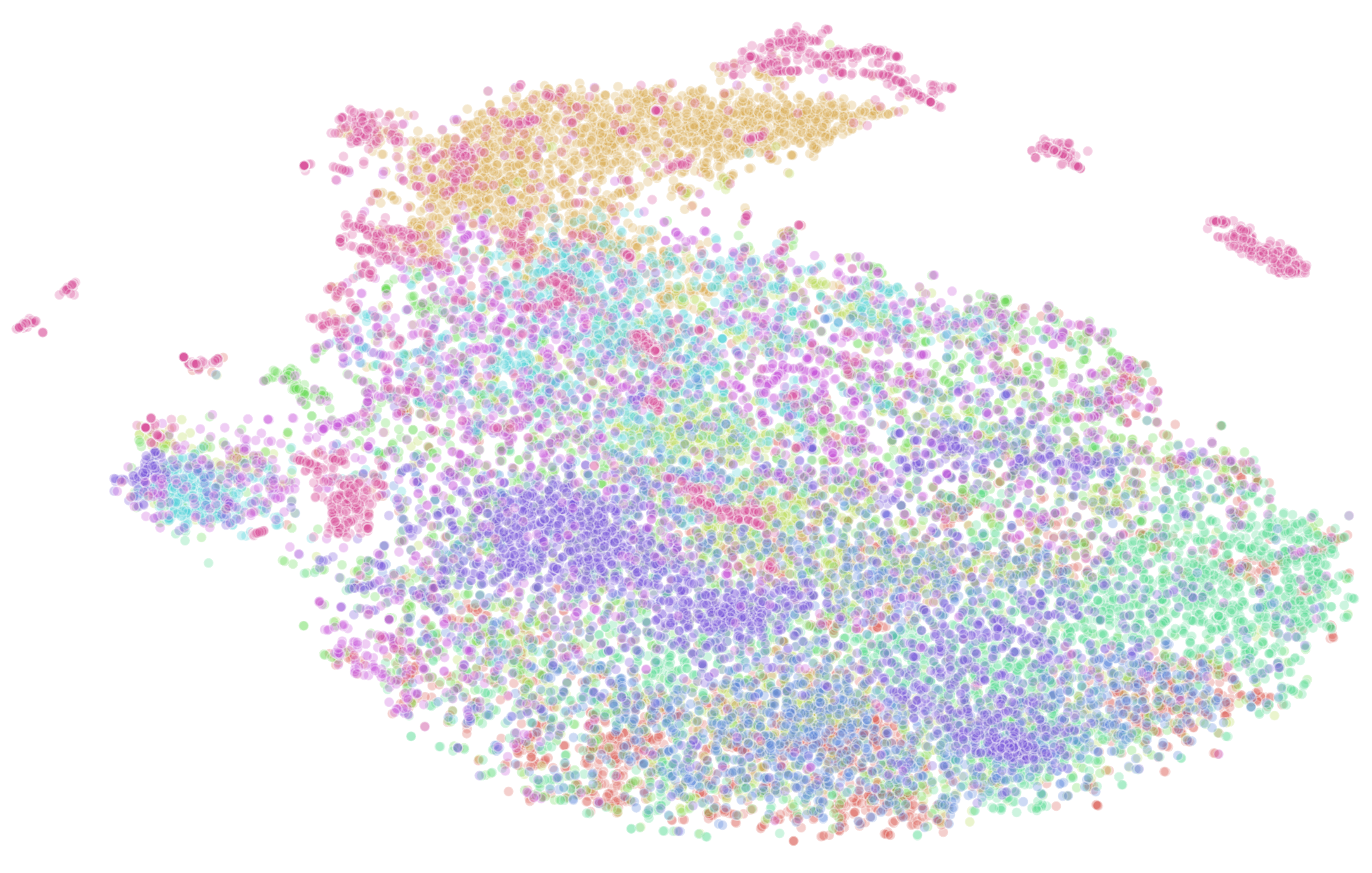}
    \includegraphics[width=0.45\linewidth]{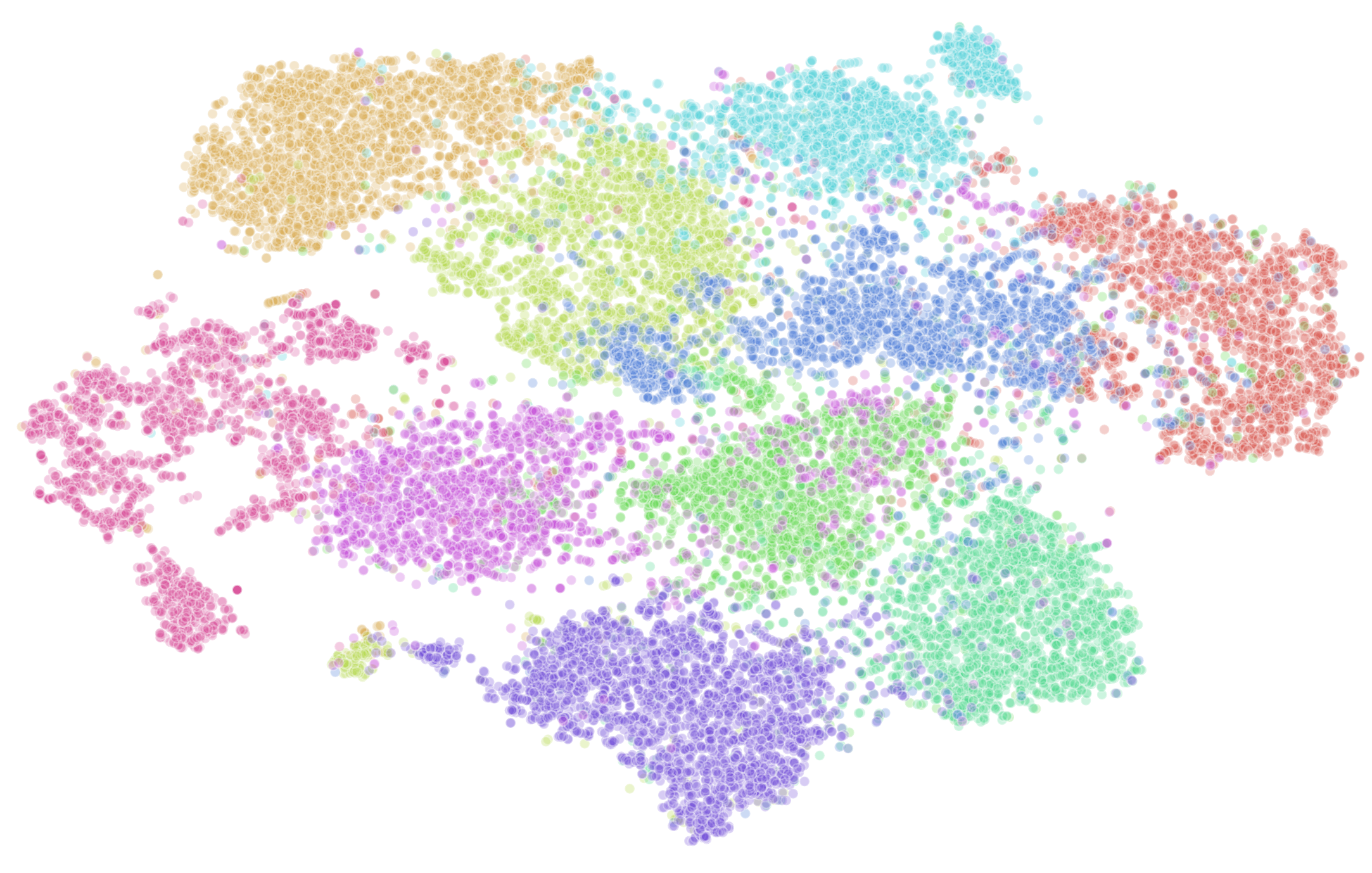}
    \caption{t-SNE visualization of EuroSAT image representations. One color represents one class. Left: random-encoded features; right: SSL-encoded features. SSL-encoded features are well clustered even without label information.}
    \label{fig:tsne}
\end{figure}

We qualitatively evaluate the data representations learned from self-supervised pre-training by visualizing the latent distributions with t-SNE (\cref{fig:tsne}). We pre-train a ResNet50 with MoCo-v2 on SSL4EO-S12, and transfer the frozen encoder to EuroSAT to calculate one 128d representation vector for each image. We then visualize all the vectors with t-SNE, and compare the distribution with a randomly initialized encoder.

\vspace{-1em}
\section{Conclusion}
\label{sec:discussion}

In this work, we present SSL4EO-S12---a large-scale multimodal, multi-temporal unlabeled dataset for self-supervised learning (SSL) in Earth observation. An extensive benchmark on various SSL methods and remote sensing applications proves the promising benefits of the proposed dataset.

SSL4EO-S12 has some limitations: 1) there's little coverage of polar regions; 2) geographical bias exists due to cloud filtering; 3) it is not strictly free of geospatial overlap; 4) medium-resolution radar and multispectral images are a limited subset of Earth observation data. Despite these, we believe SSL4EO-S12 renders a valuable basis to advance self-supervised pre-training and large-scale data mining in remote sensing.

\section*{Acknowledgments}
This work is jointly supported by the Helmholtz Association through the Framework of Helmholtz AI (grant  number:  ZT-I-PF-5-01) - Local Unit ``Munich Unit @Aeronautics, Space and Transport (MASTr)'' and Helmholtz Excellent Professorship ``Data Science in Earth Observation - Big Data Fusion for Urban Research''(grant number: W2-W3-100), by the German Federal Ministry of Education and Research (BMBF) in the framework of the international future AI lab "AI4EO -- Artificial Intelligence for Earth Observation: Reasoning, Uncertainties, Ethics and Beyond" (grant number: 01DD20001) and by German Federal Ministry for Economic Affairs and Climate Action in the framework of the "national center of excellence ML4Earth" (grant number: 50EE2201C). The computing resources were supported by the Helmholtz Association’s Initiative and Networking Fund on the HAICORE@FZJ partition.


\small{
\printbibliography[segment=0]
}


\onecolumn
\normalsize
\newrefsegment


\section*{\large\textbf{Appendix-1: Additional dataset information}}
\vspace{1em}

\subsection{\textbf{Sentinel-1/2}}
The proposed SSL4EO-S12 dataset exploits freely available SAR/optical satellite images from European Space Agency's Sentinel mission (under the CC-BY license). 

The Sentinel-1 mission \citep{torres2012gmes} consists of two polar-orbiting satellites, equipped with C-band SAR sensors, which enables them to acquire imagery regardless of the weather. For the Sentinel-1 images in the SSL4EO-S12 dataset, ground-range-detected (GRD) products with both VH and VV polarization acquired in the interferometric wide swath (IW) mode were used. These images contain the $\sigma^{0}$ backscatter coefficient in dB scale. The image resolution is 10m.

The Sentinel-2 mission \citep{drusch2012sentinel} comprises two polar-orbiting satellites in the same orbit, equipped with multi-spectral imaging sensors. For Sentinel-2 images in the SSL4EO-S12 dataset, both level-1C top-of-atmosphere reflectance (13 bands) and level-2A atmospherically corrected surface reflectance (12 bands) were included. The image resolution ranges between 10m (visible and NIR), 20m (red edge and SWIR) and 60m (aerosols).

\subsection{\textbf{Dataset statistics}} Table \ref{tab:stats-s1} and \ref{tab:stats-s2} present the mean and standard deviation of each band for each product of the proposed SSL4EO-S12 dataset.

\begin{table}[h]
\caption{Statistics of Sentinel-1 images in the SSL4EO-S12 dataset.}
\label{tab:stats-s1}
\centering

\scalebox{1.0}{
\begin{tabular}{ccc}
\hline
     & VV     & VH     \\ \hline \hline
mean & -12.59 & -20.26 \\
std  & 5.26   & 5.91   \\ \hline
\end{tabular}
}
\end{table}
\vspace{-1em}

\begin{table}[h]
\caption{Statistics of Sentinel-2 images in the SSL4EO-S12 dataset.}
\label{tab:stats-s2}
\centering
\scalebox{1.0}{

\begin{tabular}{ccccccccccccccc}
\hline
\multicolumn{2}{c}{}        & B1     & B2 (B) & B3 (G) & B4 (R) & B5     & B6     & B7     & B8     & B8A    & B9     & B10  & B11    & B12    \\ \hline \hline
\multirow{2}{*}{s2c} & mean & 1612.9 & 1397.6 & 1322.3 & 1373.1 & 1561.0 & 2108.4 & 2390.7 & 2318.7 & 2581.0 & 837.7  & 22.0 & 2195.2 & 1537.4 \\
                     & std  & 791.0  & 854.3  & 878.7  & 1144.9 & 1127.5 & 1164.2 & 1276.0 & 1249.5 & 1345.9 & 577.5  & 47.5 & 1340.0 & 1142.9 \\
\multirow{2}{*}{s2a} & mean & 756.4  & 889.6  & 1151.7 & 1307.6 & 1637.6 & 2212.6 & 2442.0 & 2538.9 & 2602.9 & 2666.8 & -    & 2388.8 & 1821.5 \\
                     & std  & 1111.4 & 1159.1 & 1188.1 & 1375.2 & 1376.6 & 1358.6 & 1418.4 & 1476.4 & 1439.9 & 1582.1 & -    & 1460.7 & 1352.2 \\ \hline
\end{tabular}

}

\end{table}

\subsection{\textbf{Data storage}} 
\begin{figure}[h]
    \centering
    \scalebox{0.5}{
    \includegraphics[width=0.6\linewidth]{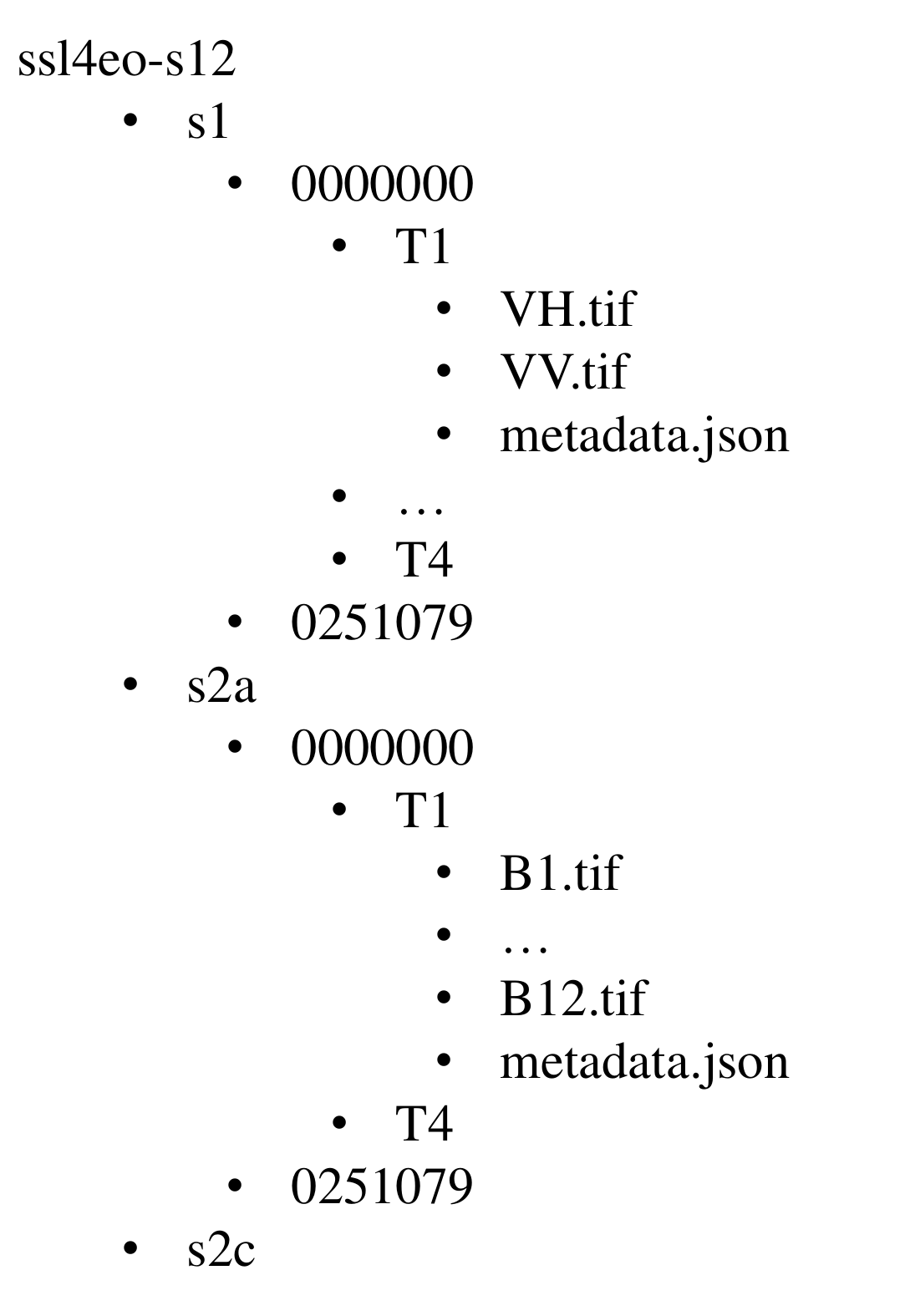}
    }
    \caption{SSL4EO-S12 file structure.}
    \label{fig:file_structure}
\end{figure}

The SSL4EO-S12 dataset is stored in GeoTiff format for each band of each patch. The file structure is shown in \cref{fig:file_structure}, where s1/s2a/s2c represents Sentinel-1 / Sentinel-2 level-2A / Sentinel-2 level-1C, and t1 - t4 represent 4 seasons. Raw files (extracted GeoTiff) occupy about 500GB/800GB/800GB disk storage for S1/S2A/S2C, and compressed tar.gz files occupy about 450GB/500GB/500GB correspondingly. If converting to uint8 and encoding with jpeg, a lossy dataset occupies less than 50 GB for each product. We later show this won't affect much the downstream performance.

\subsection{\textbf{Metadata}} Each patch comes with a metadata file that collects the image properties of this patch. See Table \ref{tab:metadata-s1} and \ref{tab:metadata-s2} for details.

\subsection{\textbf{Example visualization}} 
\begin{figure}[h]
  \centering
  \includegraphics[width=0.85\linewidth]{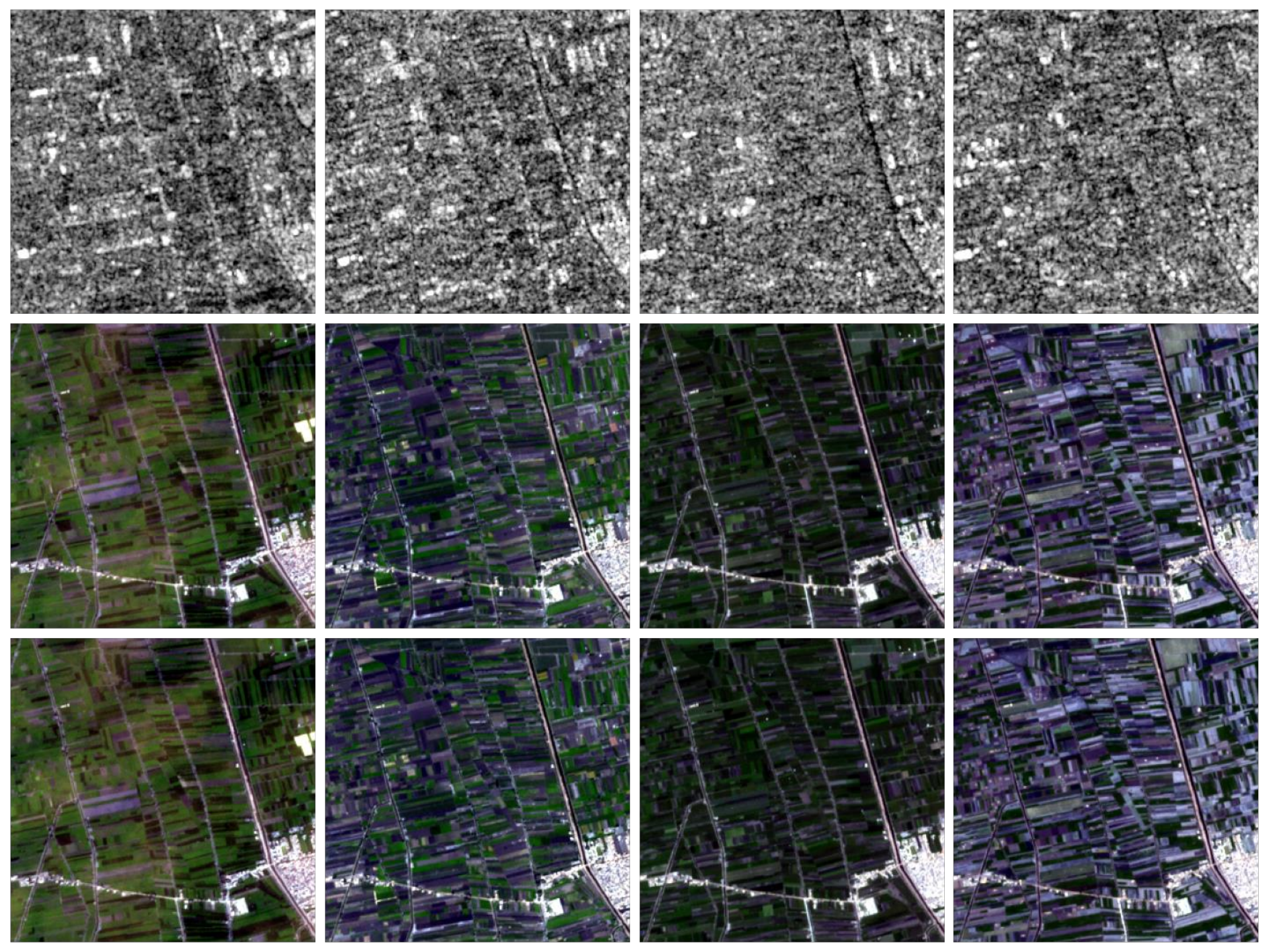}
  \caption{Sample visualization of one tile from SSL4EO-S12 dataset. The rows top-down present grayscale and false-color imagery based on the Sentinel-1 GRD product, Sentinel-2 level-1C, and Sentinel-2 level-2A multispectral data with corresponding columns representing the four seasons spring, summer, fall, and winter from left to right.}
  \label{fig:examples}

\end{figure}

Figure \ref{fig:examples} visualizes an example geospatial tile of SSL4EO-S12.

\clearpage

\section*{\large\textbf{Appendix-2: Implementation details}}
\vspace{1em}

\setcounter{subsection}{0}

\subsection{\textbf{Pre-training}}

We use Sentinel-2 level-1C images for the main pre-training experiments, which are pre-processed by converting to uint8 for efficiency (divided by 10000 and multiplied by 255, see Section \ref{subsec:pre-process}). We use 4 NVIDIA A100 GPUs with a total batch size of 256 for all the pre-training experiments. For the main experiments, we pre-train ResNet50 or Vit-S/16 for 100 epochs. Training time varies between different methods from 7 (MAE) to 25 (DINO) hours. The total experiments (including parameter tuning) take about 70k core hours (1400 GPU hours).

\vspace{0.5em}
\subsubsection{MoCo}
We pre-train the MoCo-v2/v3 models using their default settings following the publicly available repository (\url{https://github.com/facebookresearch/moco} and \url{https://github.com/facebookresearch/moco-v3}). We use RandomResizedCrop, RandomBrightness/Contrast (to have a partial color jittering for multiple bands), RandomGrayscale, RandomGaussianBlur, RandomHorizontalFlip and \texttt{RandomSeasonContrast} as data augmentations. For MoCo-v2 (ResNet50), we use SGD optimizer and cosine learning rate schedule with a learning rate 0.03. For MoCo-v3 (ViT-S/16), we use AdamW optimizer and cosine schedule with a learning rate 1.5e-4.

\vspace{0.3em}
\subsubsection{DINO}
We pre-train the DINO models using its default settings following the publicly available repository (\url{https://github.com/facebookresearch/dino}). The data augmentations include those of MoCo, as well as additional Multi-Crop and Solarization. For ResNet50, we use SGD optimizer and cosine learning rate schedule with a learning rate 0.03. For ViT-S/16, we use AdamW optimizer and cosine learning rate schedule with a learning rate 1.5e-4.

\vspace{0.3em}
\subsubsection{MAE}
We pre-train the MAE models using its default settings following the publicly available repository (\url{https://github.com/facebookresearch/mae}). The mask ratio is set to 0.7. The data augmentations include RandomResizedCrop, RandomHorizontalFlip and RandomSeason. We use AdamW optimizer and cosine learning rate schedule with a learning rate 1.5e-4.

\vspace{0.3em}
\subsubsection{data2vec}
We pre-train the data2vec models using its default settings following the publicly available repository (\url{https://github.com/facebookresearch/fairseq/tree/main/examples/data2vec}). The data augmentations include Resize/CenterCrop, RandomHorizontalFlip and RandomSeason. We use AdamW optimizer and cosine learning rate schedule with a learning rate 1e-3.

\subsection{\textbf{Downstream tasks}} Below are additional implementation details for the downstream tasks.

\vspace{0.5em}
\subsubsection{EuroSAT}
We split EuroSAT into 21600 training and 5400 testing images for evaluation. The data augmentations are RandomResizedCrop/RandomHorizontalFlip for training, and Resize/CenterCrop for testing. We resize the images to 224x224 for better performance (see Section \ref{subsec:pre-process}). The batch size is 256. We use CrossEntropyLoss and SGD optimizer with step decay learning rate (divided by 10 at epoch 60 and 80) for 100 epochs. We use simple grid search strategy to find suitable learning rates for linear probing and fine-tuning.

\vspace{0.3em}
\subsubsection{BigEarthNet}
We use 311667 training and 103944 testing images from BigEarthNet for evaluation. Different settings of the amount of labels affect only the training split. The data augmentations are RandomResizedCrop/RandomHorizontalFlip for training, and Resize/CenterCrop for testing. We use a cropping scale of 0.8 to avoid strong occlusions (BigEarthNet is a multi-label dataset). We resize the images to 224x224 for better performance. The batch size is 256. We use MultiLabelSoftMarginLoss and SGD optimizer with step decay learning rate (divided by 10 at epochs 60 and 80) for 100 epochs. We use a simple grid search strategy to find suitable learning rates for linear probing and fine-tuning.

\vspace{0.3em}
\subsubsection{So2Sat-LCZ42}
We use 352366 training and 24119 testing Sentinel-2 images from So2Sat-LCZ42 for evaluation. Different settings of the amount of labels affect only the training split. We use the \textit{culture-10} version of So2Sat-LCZ42: the training data and the testing data are from different cities. The data augmentations are RandomResizedCrop/RandomHorizontalFlip for training, and Resize/CenterCrop for testing. We resize the images to 224x224 for better performance. The batch size is 256. We use CrossEntropyLoss and SGD optimizer with step decay learning rate (divided by 10 at epochs 60 and 80) for 100 epochs. We use a simple grid search strategy to find suitable learning rates for linear probing and fine-tuning.

\vspace{0.3em}
\subsubsection{DFC2020}
We use 5128 training and 986 testing Sentinel-2 images from DFC2020 dataset for evaluation. The batch size is set to 8 and we train the models for 50 epochs. We use CrossEntropyLoss and SGD optimizer with momentum 0.9 and weight decay 5e-4. The initial learning rate is 1e-3, which is decayed by a factor of 0.9 in every epoch until 1e-4. 

\vspace{0.3em}
\subsubsection{OSCD} 
This dataset is composed of 24 pairs of multispectral images from Sentinel-2 in total. Following \citep{daudt2018urban}, we use 14 of them for training, and the rest for testing. In fine-tuning stage, we adopt settings similar to those in \citep{manas2021seasonal}. That is, the original images are split into non-overlapping patches of 96 × 96 pixels as inputs, which leads to 827 and 285 patches for training and testing, respectively. The batch size is set to 32, and we in total train 50 epochs. We use Adam optimizer with a weight decay of 1e-4. The initial learning rate is 1e-3, and decreases exponentially with a multiplicative factor of 0.95 for every epoch. The resulting models are evaluated on the whole test set for overall precision, recall and F1 score (with a default threshold 0.5). 

\clearpage
\section*{\large\textbf{Appendix-3: Additional experimental results}}
\vspace{1em}

\label{subsec:pre-process}
\setcounter{subsection}{0}

\subsection{\textbf{Effect of data pre-processing}}
We show the influence of data pre-processing (for pre-training) in Table \ref{tab:pre-process}, where int16 means 16 bits raw input, uint8 means compressed 8 bits input (divided by 10000 and multiplied by 255), uint8-n means normalization by mean and standard deviation, and L2A/L1C means with/without atmospheric correction. The results show similar performance between 16 bit and 8 bit, supporting compressed input as it saves a lot of storage space and computing time. The results also show comparable performance for L1C and L2A, as well as for the use of normalization. Therefore, in our main experiments, we use level-1C, uint8, unnormalized data for pre-training.

\begin{table}[h!]
\centering
\caption{Effects of pre-processing for pretraining. We pre-train a ResNet50 with MoCo-v2 on a 50k subset for efficiency and report linear probing results.}

\begin{tabular}{ccccccc}
\hline
         & \multicolumn{3}{c}{BE-10\% (L2A)} & \multicolumn{3}{c}{EuroSAT (L1C)} \\
         & int16  & uint8  & uint8-n  & int16   & uint8  & uint8-n  \\ \hline \hline
L2A & 73.9   & 73.9   & 75.6     & 85.2    & 86.6   & -        \\
L1C & 73.8   & 74     & -        & 86.2    & 87.7   & 85.1     \\ \hline
\end{tabular}
\label{tab:pre-process}
\end{table}

\subsection{\textbf{Effect of input image size}}
We analyze the impact of image resolution on pre-training and transfer learning in Table \ref{tab:input-size}. We clearly observe the advantage of upsampling the input image size. Therefore, in our main experiments, we upscale the downstream input images to 224x224 for better performance.

\begin{table}[h!]
\centering
\caption{Effects of input size of downstream tasks. We pre-train ResNet50 with MoCo-v2 on the 50k subset for efficiency and report linear probing results.}
\begin{tabular}{ccccccc}
\hline
             & \multicolumn{3}{c}{BE-10\%} & \multicolumn{3}{c}{EuroSAT}                      \\
             & 224     & \textbf{112}     & 56      & 224                & 112  & \textbf{56} \\ \hline \hline
pretrain 224 & \textbf{75.1}             & 71.1    & -       & \textbf{93.1} & 89.4 & 89.4        \\
pretrain 112 & \textbf{75.9}             & 73.9    & -       & \textbf{92.7} & 91.1 & 86.6        \\
pretrain 56  & 73.1             & 70.5    & 70.5    & 89.7                        & 89   & 86.4        \\ \hline
\end{tabular}
\label{tab:input-size}
\end{table}

\subsection{\textbf{Effect of MAE masking ratio}} Table \ref{tab:mask-ratio} shows the influence of masking ratios in MAE during pre-training. We find 70\% to be the best masking ratio, which is similar to natural images as reported in MAE paper, where 75\% is the best. It is also promising to see that the model still learns good representations even with 90\% pixels masked. 

\begin{table}[h!]
\centering
\caption{Effects of different masking ratios. We pre-train ViT-S/16 with MAE on the 50k subset for efficiency and report linear probing results on BigEarthNet-10\%.}
\begin{tabular}{ccccccc}
\hline
Mask ratio & 90\% & 80\%  & 75\%  & 70\%         & 60\%  & 50\%  \\ \hline \hline
BE-10\% & 72  & 73.5 & 73.8 & \textbf{74} & 73.6 & 72.9 \\ \hline
\end{tabular}
\label{tab:mask-ratio}
\end{table}

\subsection{\textbf{Effect of different pre-training protocols}}

\subsubsection{Different ImageNet pre-training protocols} \cref{tab:imagenet-diff} shows a comparison of different ImageNet pre-training protocols. We pre-train ResNet50 with MoCo-v2 for self-supervised pre-training, and report fine-tuning results on BigEarthNet. The table shows that ImageNet pre-training provides good representations that can be generalized well to remote sensing images with RGB but not all bands. It can also be seen that when using RGB, self-supervised pre-training on ImageNet can further improve the downstream performance in remote sensing compared to supervised pre-training.

\begin{table}[h!]
\caption{Summary of different ImageNet pre-training protocols.}
\label{tab:imagenet-diff}
\centering
\begin{tabular}{ccccc}
\hline
downstr./               & \multicolumn{2}{c}{All optical bands} & \multicolumn{2}{c}{RGB}       \\
pre-train               & BE-10\%           & BE-100\%          & BE-10\%       & BE-100\%      \\ \hline \hline
Supervised     & 83.4              & 88.7              & 69.5          & 79.0          \\
ImageNet (Sup) & 82.5              & 89.5              & 80.0          & 86.7          \\
ImageNet (SSL) & 82.5              & 89.4              & 81.9          & 90.8          \\
SSL4EO-S12 (SSL)     & \textbf{86.2}     & \textbf{91.8}     & \textbf{82.7} & \textbf{90.9} \\ \hline
\end{tabular}
\end{table}

\subsubsection{\textbf{Supervised pre-training on RS datasets}}
\cref{tab:rs-sup} shows a comparison of supervised and unsupervised pre-training on remote sensing datasets. We do supervised pre-training on BigEarthNet and self-supervised pre-training (MoCo-v2, ResNet50) on both BigEarthNet and SSL4EO-S12. We evaluate the pre-trained models on EuroSAT. The results show that self-supervised pre-training outperforms supervised pre-training on remote sensing data.

\begin{table}[h!]
\centering
\caption{A comparison of different pre-training protocols on remote sensing datasets and evaluated on EuroSAT.}
\label{tab:rs-sup}
\begin{tabular}{ccccc}
\hline
downstr./                 & \multicolumn{2}{c}{Linear probing} & \multicolumn{2}{c}{Fine-tuning} \\
pre-train                   & 10\%     & 100\%   & 10\%   & 100\%  \\ \hline \hline
BigEarthNet (Sup) & 80.3             & 89.3            & 94.2           & 98.7           \\
BigEarthNet (SSL) & 90.7             & 94.4            & 96.3           & 98.9           \\
SSL4EO-S12 (SSL)  & \textbf{92.7}    & \textbf{98.0}   & \textbf{96.9}  & \textbf{99.1}  \\ \hline
\end{tabular}
\end{table}

\subsection{\textbf{Additional dataset comparison results}}

\Cref{tab:dataset-comp-ft} reports fine-tuning results pre-trained on different datasets in complement to the main paper. A difference most notable is ImageNet's catch-up in performance compared to the other geospatial pre-training datasets. However, a $\sim5\%$ margin when compared to SSL4EO-S12 persists. The observed is a characteristic feature of the data \textit{domain gap}: while pre-training on ImageNet learns good representations, the weights' distribution is shifted towards natural images, which can be further adjusted to remote sensing data with fine-tuning. We note that fine-tuning is more computationally expensive compared to linear probing, and \cref{tab:dataset-comp-ft} demonstrate: pre-training on SSL4EO-S12 outperforms all other datasets for downstream classification.

\begin{table}[h]
\centering
\caption{Comparison of different pre-training datasets under fine-tuning evaluation.}\label{tab:dataset-comp-ft}
\scalebox{0.9}{
\begin{tabular}{lccccc}
\hline
 dataset   & EuroSAT      & BE-10\%       & BE-100\%      & So2Sat-10\%   & So2Sat-100\%    \\ \hline\hline
\it ImageNet (RGB)     &\it  96.5                 & \it 80.0          & \it 86.7          & -             & -              \\
\it SeCo (RGB)           & -     & \it 80.2          & \it 86.1          & -             & -              \\
SSL4EO-S12 (RGB)    & \textbf{98.0}                                        & \textbf{82.7} & \textbf{90.9} & -             & -                 \\
BigEarthNet (MS)           & 98.9                             & 85.5          & 89.3          & 53.0          & 53.0         \\
SSL4EO-S12 (MS)            & \textbf{99.1}                  & \textbf{86.2} & \textbf{91.8} & \textbf{60.4} & \textbf{60.9}  \\ \hline
\end{tabular}
}
\end{table}

\newpage
\section*{\large\textbf{Appendix-4: Metadata}}

\begin{table*}[h]
\caption{Metadata Sentinel-1 GRD\cite{gorelick2017google}.}
\label{tab:metadata-s1}
\centering
\scalebox{0.75}{
\begin{tabular}{llp{12cm}}
\hline
\multicolumn{1}{c}{\textbf{Name}}                & \multicolumn{1}{c}{\textbf{Type}} & \multicolumn{1}{c}{\textbf{Description}}                                                                                                                                \\ \hline
GRD\_Post\_Processing\_facility\_country         & STRING                            & Name of the country where the facility is located. This element is configurable within the IPF.                                                                         \\
GRD\_Post\_Processing\_facility\_name            & STRING                            & Name of the facility where the processing step was performed. This element is configurable within the IPF.                                                              \\
GRD\_Post\_Processing\_facility\_organisation    & STRING                            & Name of the organisation responsible for the facility. This element is configurable within the IPF.                                                                     \\
GRD\_Post\_Processing\_facility\_site            & STRING                            & Geographical location of the facility. This element is configurable within the IPF.                                                                                     \\
GRD\_Post\_Processing\_software\_name            & STRING                            & Name of the software.                                                                                                                                                   \\
GRD\_Post\_Processing\_software\_version         & STRING                            & Software version identification.                                                                                                                                        \\
GRD\_Post\_Processing\_start                     & DOUBLE                            & Processing start time.                                                                                                                                                  \\
GRD\_Post\_Processing\_stop                      & DOUBLE                            & Processing stop time.                                                                                                                                                   \\
SLC\_Processing\_facility\_country               & STRING                            & Name of the country where the facility is located. This element is configurable within the IPF.                                                                         \\
SLC\_Processing\_facility\_name                  & STRING                            & Name of the facility where the processing step was performed. This element is configurable within the IPF.                                                              \\
SLC\_Processing\_facility\_organisation          & STRING                            & Name of the organisation responsible for the facility. This element is configurable within the IPF.                                                                     \\
SLC\_Processing\_facility\_site                  & STRING                            & Geographical location of the facility. This element is configurable within the IPF.                                                                                     \\
SLC\_Processing\_software\_name                  & STRING                            & Name of the software.                                                                                                                                                   \\
SLC\_Processing\_software\_version               & STRING                            & Software version identification.                                                                                                                                        \\
SLC\_Processing\_start                           & DOUBLE                            & Processing start time.                                                                                                                                                  \\
SLC\_Processing\_stop                            & DOUBLE                            & Processing stop time.                                                                                                                                                   \\
S1TBX\_Calibration\_Operator\_version            & STRING                            & Sentinel-1 Toolbox calibration tool version.                                                                                                                            \\
S1TBX\_SAR\_Processing\_version                  & STRING                            & Sentinel-1 Toolbox SAR processing tool version.                                                                                                                         \\
SNAP\_Graph\_Processing\_Framework\_GPF\_version & STRING                            & Sentinel Application Platform (SNAP) version.                                                                                                                           \\
startTimeANX                                     & DOUBLE                            & Sensing start time of the input data relative to the ascending node crossing. This is a count of the time elapsed since the orbit ascending node crossing {[}ms{]}.     \\
stopTimeANX                                      & DOUBLE                            & Sensing stop time of the input data relative to the ascending node crossing. This is a count of the time elapsed since the orbit ascending node crossing {[}ms{]}.      \\
nssdcIdentifier                                  & STRING                            & Uniquely identifies the mission according to standards defined by the World Data Center for Satellite Information (WDC-SI), available here.                             \\
familyName                                       & STRING                            & The full mission name. E.g. “SENTINEL-1”                                                                                                                                \\
platform\_number                                 & STRING                            & The alphanumeric identifier of the platform within the mission.                                                                                                         \\
instrument                                       & STRING                            & Information related to the instrument on the platform to which acquired the data.                                                                                       \\
instrumentMode                                   & STRING                            & IW (Interferometric Wide Swath), EW (Extra Wide Swath) or SM (Strip Map)                                                                                                \\
instrumentSwath                                  & STRING                            & List of the swaths contained within a product. Most products will contain only one swath, except for TOPS SLC products which include 3 or 5 swaths.                     \\
orbitNumber\_start                               & DOUBLE                            & Absolute orbit number of the oldest line within the image data.                                                                                                         \\
orbitNumber\_stop                                & DOUBLE                            & Absolute orbit number of the most recent line within the image data.                                                                                                    \\
relativeOrbitNumber\_start                       & DOUBLE                            & Relative orbit number of the oldest line within the image data.                                                                                                         \\
relativeOrbitNumber\_stop                        & DOUBLE                            & Relative orbit number of the most recent line within the image data.                                                                                                    \\
cycleNumber                                      & DOUBLE                            & Absolute sequence number of the mission cycle to which the oldest image data applies.                                                                                   \\
phaseIdentifier                                  & DOUBLE                            & Id of the mission phase to which the oldest image data applies.                                                                                                         \\
orbitProperties\_pass                            & STRING                            & Direction of the orbit ('ASCENDING' or 'DESCENDING') for the oldest image data in the product (the start of the product).                                               \\
orbitProperties\_ascendingNodeTime               & DOUBLE                            & UTC time of the ascending node of the orbit. This element is present for all products except ASAR L2 OCN products which are generated from an ASAR L1 input.            \\
resolution                                       & STRING                            & H for high or M for medium.                                                                                                                                             \\
resolution\_meters                               & DOUBLE                            & Resolution in meters.                                                                                                                                                   \\
instrumentConfigurationID                        & DOUBLE                            & The instrument configuration ID (Radar database ID) for this data.                                                                                                      \\
missionDataTakeID                                & DOUBLE                            & Unique ID of the datatake within the mission.                                                                                                                           \\
transmitterReceiverPolarisation                  & DOUBLE                            & Transmit/Receive polarisation for the data. There is one element for each Tx/Rx combination: {[}''VV''{]}, {[}''HH''{]}, {[}''VV'', ''VH''{]}, or {[}''HH'', ''HV''{]}. \\
productClass                                     & STRING                            & Output product class “A” for Annotation or “S” for Standard.                                                                                                            \\
productClassDescription                          & STRING                            & Textual description of the output product class.                                                                                                                        \\
productComposition                               & STRING                            & The composition type of this product: "Individual", "Slice", or "Assembled".                                                                                            \\
productType                                      & STRING                            & The product type (correction level) of this product.                                                                                                                    \\
productTimelinessCategory                        & STRING                            & Describes the required timeliness of the processing. One of: NRT-10m, NRT-1h, NRT-3h, Fast-24h, Off-line, or Reprocessing                                               \\
sliceProductFlag                                 & STRING                            & True if this is a slice from a larger product or false if this is a complete product.                                                                                   \\
segmentStartTime                                 & DOUBLE                            & Sensing start time of the segment to which this slice belongs. This field is only present if sliceProductFlag = true                                                    \\
sliceNumber                                      & DOUBLE                            & Absolute slice number of this slice starting at 1. This field is only present if sliceProductFlag = true.                                                               \\
totalSlices                                      & DOUBLE                            & Total number of slices in the complete data take. This field is only present if sliceProductFlag = true.                                                                \\ \hline
\end{tabular}
}
\end{table*}

\begin{table*}[t]
\caption{Metadata Sentinel-2 L1C\cite{gorelick2017google}.}
\label{tab:metadata-s2}
\centering
\scalebox{0.7}{
\begin{tabular}{llp{14cm}}
\hline
\multicolumn{1}{c}{\textbf{Name}}       & \multicolumn{1}{c}{\textbf{Type}} & \multicolumn{1}{c}{\textbf{Description}}                                                                                                                             \\ \hline
AOT\_RETRIEVAL\_ACCURACY                & DOUBLE                            & Accuracy of Aerosol Optical thickness model                                                                                                                          \\
CLOUDY\_PIXEL\_PERCENTAGE               & DOUBLE                            & Granule-specific cloudy pixel percentage taken from the original metadata                                                                                            \\
CLOUD\_COVERAGE\_ASSESSMENT             & DOUBLE                            & Cloudy pixel percentage for the whole archive that contains this granule. Taken from the original metadata                                                           \\
CLOUDY\_SHADOW\_PERCENTAGE              & DOUBLE                            & Percentage of pixels classified as cloud shadow                                                                                                                      \\
DARK\_FEATURES\_PERCENTAGE              & DOUBLE                            & Percentage of pixels classified as dark features or shadows                                                                                                          \\
DATASTRIP\_ID                           & STRING                            & Unique identifier of the datastrip Product Data Item (PDI)                                                                                                           \\
DATATAKE\_IDENTIFIER                    & STRING                            & Uniquely identifies a given Datatake. The ID contains the Sentinel-2 satellite, start date and time, absolute orbit number, and processing baseline.                 \\
DATATAKE\_TYPE                          & STRING                            & MSI operation mode                                                                                                                                                   \\
DEGRADED\_MSI\_DATA\_PERCENTAGE         & DOUBLE                            & Percentage of degraded MSI and ancillary data                                                                                                                        \\
FORMAT\_CORRECTNESS                     & STRING                            & Synthesis of the On-Line Quality Control (OLQC) checks performed at granule (Product\_Syntax) and datastrip (Product Syntax and DS\_Consistency) levels              \\
GENERAL\_QUALITY                        & STRING                            & Synthesis of the OLQC checks performed at the datastrip level (Relative\_Orbit\_Number)                                                                              \\
GENERATION\_TIME                        & DOUBLE                            & Product generation time                                                                                                                                              \\
GEOMETRIC\_QUALITY                      & STRING                            & Synthesis of the OLQC checks performed at the datastrip level (Attitude\_Quality\_Indicator)                                                                         \\
GRANULE\_ID                             & STRING                            & Unique identifier of the granule PDI (PDI\_ID)                                                                                                                       \\
HIGH\_PROBA\_CLOUDS\_PERCENTAGE         & DOUBLE                            & Percentage of pixels classified as high probability clouds                                                                                                           \\
MEAN\_INCIDENCE\_AZIMUTH\_ANGLE\_B1     & DOUBLE                            & Mean value containing viewing incidence azimuth angle average for band B1 and for all detectors                                                                      \\
MEAN\_INCIDENCE\_AZIMUTH\_ANGLE\_B2     & DOUBLE                            & Mean value containing viewing incidence azimuth angle average for band B2 and for all detectors                                                                      \\
MEAN\_INCIDENCE\_AZIMUTH\_ANGLE\_B3     & DOUBLE                            & Mean value containing viewing incidence azimuth angle average for band B3 and for all detectors                                                                      \\
MEAN\_INCIDENCE\_AZIMUTH\_ANGLE\_B4     & DOUBLE                            & Mean value containing viewing incidence azimuth angle average for band B4 and for all detectors                                                                      \\
MEAN\_INCIDENCE\_AZIMUTH\_ANGLE\_B5     & DOUBLE                            & Mean value containing viewing incidence azimuth angle average for band B5 and for all detectors                                                                      \\
MEAN\_INCIDENCE\_AZIMUTH\_ANGLE\_B6     & DOUBLE                            & Mean value containing viewing incidence azimuth angle average for band B6 and for all detectors                                                                      \\
MEAN\_INCIDENCE\_AZIMUTH\_ANGLE\_B7     & DOUBLE                            & Mean value containing viewing incidence azimuth angle average for band B7 and for all detectors                                                                      \\
MEAN\_INCIDENCE\_AZIMUTH\_ANGLE\_B8     & DOUBLE                            & Mean value containing viewing incidence azimuth angle average for band B8 and for all detectors                                                                      \\
MEAN\_INCIDENCE\_AZIMUTH\_ANGLE\_B8A    & DOUBLE                            & Mean value containing viewing incidence azimuth angle average for band B8a and for all detectors                                                                     \\
MEAN\_INCIDENCE\_AZIMUTH\_ANGLE\_B9     & DOUBLE                            & Mean value containing viewing incidence azimuth angle average for band B9 and for all detectors                                                                      \\
MEAN\_INCIDENCE\_AZIMUTH\_ANGLE\_B10    & DOUBLE                            & Mean value containing viewing incidence azimuth angle average for band B10 and for all detectors                                                                     \\
MEAN\_INCIDENCE\_AZIMUTH\_ANGLE\_B11    & DOUBLE                            & Mean value containing viewing incidence azimuth angle average for band B11 and for all detectors                                                                     \\
MEAN\_INCIDENCE\_AZIMUTH\_ANGLE\_B12    & DOUBLE                            & Mean value containing viewing incidence azimuth angle average for band B12 and for all detectors                                                                     \\
MEAN\_INCIDENCE\_ZENITH\_ANGLE\_B1      & DOUBLE                            & Mean value containing viewing incidence zenith angle average for band B1 and for all detectors                                                                       \\
MEAN\_INCIDENCE\_ZENITH\_ANGLE\_B2      & DOUBLE                            & Mean value containing viewing incidence zenith angle average for band B2 and for all detectors                                                                       \\
MEAN\_INCIDENCE\_ZENITH\_ANGLE\_B3      & DOUBLE                            & Mean value containing viewing incidence zenith angle average for band B3 and for all detectors                                                                       \\
MEAN\_INCIDENCE\_ZENITH\_ANGLE\_B4      & DOUBLE                            & Mean value containing viewing incidence zenith angle average for band B4 and for all detectors                                                                       \\
MEAN\_INCIDENCE\_ZENITH\_ANGLE\_B5      & DOUBLE                            & Mean value containing viewing incidence zenith angle average for band B5 and for all detectors                                                                       \\
MEAN\_INCIDENCE\_ZENITH\_ANGLE\_B6      & DOUBLE                            & Mean value containing viewing incidence zenith angle average for band B6 and for all detectors                                                                       \\
MEAN\_INCIDENCE\_ZENITH\_ANGLE\_B7      & DOUBLE                            & Mean value containing viewing incidence zenith angle average for band B7 and for all detectors                                                                       \\
MEAN\_INCIDENCE\_ZENITH\_ANGLE\_B8      & DOUBLE                            & Mean value containing viewing incidence zenith angle average for band B8 and for all detectors                                                                       \\
MEAN\_INCIDENCE\_ZENITH\_ANGLE\_B8A     & DOUBLE                            & Mean value containing viewing incidence zenith angle average for band B8a and for all detectors                                                                      \\
MEAN\_INCIDENCE\_ZENITH\_ANGLE\_B9      & DOUBLE                            & Mean value containing viewing incidence zenith angle average for band B9 and for all detectors                                                                       \\
MEAN\_INCIDENCE\_ZENITH\_ANGLE\_B10     & DOUBLE                            & Mean value containing viewing incidence zenith angle average for band B10 and for all detectors                                                                      \\
MEAN\_INCIDENCE\_ZENITH\_ANGLE\_B11     & DOUBLE                            & Mean value containing viewing incidence zenith angle average for band B11 and for all detectors                                                                      \\
MEAN\_INCIDENCE\_ZENITH\_ANGLE\_B12     & DOUBLE                            & Mean value containing viewing incidence zenith angle average for band B12 and for all detectors                                                                      \\
MEAN\_SOLAR\_AZIMUTH\_ANGLE             & DOUBLE                            & Mean value containing sun azimuth angle average for all bands and detectors                                                                                          \\
MEAN\_SOLAR\_ZENITH\_ANGLE              & DOUBLE                            & Mean value containing sun zenith angle average for all bands and detectors                                                                                           \\
MEDIUM\_PROBA\_CLOUDS\_PERCENTAGE       & DOUBLE                            & Percentage of pixels classified as medium probability clouds                                                                                                         \\
MGRS\_TILE                              & STRING                            & US-Military Grid Reference System (MGRS) tile                                                                                                                        \\
NODATA\_PIXEL\_PERCENTAGE               & DOUBLE                            & Percentage of No Data pixels                                                                                                                                         \\
NOT\_VEGETATED\_PERCENTAGE              & DOUBLE                            & Percentage of pixels classified as non-vegetated                                                                                                                     \\
PROCESSING\_BASELINE                    & STRING                            & Configuration baseline used at the time of the product generation in terms of processor software version and major Ground Image Processing Parameters (GIPP) version \\
PRODUCT\_ID                             & STRING                            & The full id of the original Sentinel-2 product                                                                                                                       \\
RADIATIVE\_TRANSFER\_ACCURACY           & DOUBLE                            & Accuracy of radiative transfer model                                                                                                                                 \\
RADIOMETRIC\_QUALITY                    & STRING                            & Based on the OLQC reports contained in the Datastrips/QI\_DATA with RADIOMETRIC\_QUALITY checklist name                                                              \\
REFLECTANCE\_CONVERSION\_CORRECTION     & DOUBLE                            & Earth-Sun distance correction factor                                                                                                                                 \\
SATURATED\_DEFECTIVE\_PIXEL\_PERCENTAGE & DOUBLE                            & Percentage of saturated or defective pixels                                                                                                                          \\
SENSING\_ORBIT\_DIRECTION               & STRING                            & Imaging orbit direction                                                                                                                                              \\
SENSING\_ORBIT\_NUMBER                  & DOUBLE                            & Imaging orbit number                                                                                                                                                 \\
SENSOR\_QUALITY                         & STRING                            & Synthesis of the OLQC checks performed at granule (Missing\_Lines, Corrupted\_ISP, and Sensing\_Time) and datastrip (Degraded\_SAD and Datation\_Model) levels       \\
SOLAR\_IRRADIANCE\_B1                   & DOUBLE                            & Mean solar exoatmospheric irradiance for band B1                                                                                                                     \\
SOLAR\_IRRADIANCE\_B2                   & DOUBLE                            & Mean solar exoatmospheric irradiance for band B2                                                                                                                     \\
SOLAR\_IRRADIANCE\_B3                   & DOUBLE                            & Mean solar exoatmospheric irradiance for band B3                                                                                                                     \\
SOLAR\_IRRADIANCE\_B4                   & DOUBLE                            & Mean solar exoatmospheric irradiance for band B4                                                                                                                     \\
SOLAR\_IRRADIANCE\_B5                   & DOUBLE                            & Mean solar exoatmospheric irradiance for band B5                                                                                                                     \\
SOLAR\_IRRADIANCE\_B6                   & DOUBLE                            & Mean solar exoatmospheric irradiance for band B6                                                                                                                     \\
SOLAR\_IRRADIANCE\_B7                   & DOUBLE                            & Mean solar exoatmospheric irradiance for band B7                                                                                                                     \\
SOLAR\_IRRADIANCE\_B8                   & DOUBLE                            & Mean solar exoatmospheric irradiance for band B8                                                                                                                     \\
SOLAR\_IRRADIANCE\_B8A                  & DOUBLE                            & Mean solar exoatmospheric irradiance for band B8a                                                                                                                    \\
SOLAR\_IRRADIANCE\_B9                   & DOUBLE                            & Mean solar exoatmospheric irradiance for band B9                                                                                                                     \\
SOLAR\_IRRADIANCE\_B10                  & DOUBLE                            & Mean solar exoatmospheric irradiance for band B10                                                                                                                    \\
SOLAR\_IRRADIANCE\_B11                  & DOUBLE                            & Mean solar exoatmospheric irradiance for band B11                                                                                                                    \\
SOLAR\_IRRADIANCE\_B12                  & DOUBLE                            & Mean solar exoatmospheric irradiance for band B12                                                                                                                    \\
SNOW\_ICE\_PERCENTAGE                   & DOUBLE                            & Percentage of pixels classified as snow or ice                                                                                                                       \\
SPACECRAFT\_NAME                        & STRING                            & Sentinel-2 spacecraft name: Sentinel-2A, Sentinel-2B                                                                                                                 \\
THIN\_CIRRUS\_PERCENTAGE                & DOUBLE                            & Percentage of pixels classified as thin cirrus clouds                                                                                                                \\
UNCLASSIFIED\_PERCENTAGE                & DOUBLE                            & Percentage of unclassified pixels                                                                                                                                    \\
VEGETATION\_PERCENTAGE                  & DOUBLE                            & Percentage of pixels classified as vegetation                                                                                                                        \\
WATER\_PERCENTAGE                       & DOUBLE                            & Percentage of pixels classified as water                                                                                                                             \\
WATER\_VAPOUR\_RETRIEVAL\_ACCURACY      & DOUBLE                            & Declared accuracy of the Water Vapor model                                                                                                                           \\ \hline
\end{tabular}
}
\end{table*}

\clearpage
\section*{\large\textbf{Datasheets for Datasets}}
\setcounter{subsection}{0}

\vspace{1em}
Here we answer the questions outlined in the datasheets for datasets paper by Gebru et al. \cite{gebru2021datasheets}.

\subsection{Motivation}

\textbf{For what purpose was the dataset created?} The dataset was created for unsupervised pre-training in Earth observation. By integrating global coverage, multiple modalities and multiple timestamps, the dataset is intended to serve for diverse applications in remote sensing. The dataset fills the gap between multiple existing pre-training datasets, e.g. domain gap of ImageNet, regional coverage of BigEarthNet \cite{sumbul2019bigearthnet}, single modality of SeCo \cite{manas2021seasonal}, single timestamp of SEN12MS \cite{Schmitt2019}, and patch overlap of both SEN12MS and SeCo.

\textbf{Who created the dataset (e.g., which team, research group) and on behalf of which entity (e.g.,
company, institution, organization)?} The dataset was created by the lab "Data Science in Earth Observation" at Technical University of Munich and German Aerospace Center.

\textbf{Who funded the creation of the dataset?} The creation of the dataset was funded by the Helmholtz Association through the Framework of Helmholtz AI (grant number: ZT-I-PF-5-01) - Local Unit “Munich Unit @Aeronautics, Space and Transport (MASTr)”. The computing resources for benchmark experiments were supported by the Helmholtz Association's Initiative and Networking Fund on the HAICORE@FZJ partition.

\subsection{Composition}

\textbf{What do the instances that comprise the dataset represent (e.g., documents, photos, people, countries)?} This dataset only contains satellite images. In addition we provide meta-data for these images, which contain information about data acquisition.

\textbf{How many instances are there in total (of each type, if appropriate)?} The dataset contains 251079 geographical patches, each patch including 3 product types and 4 seasons. In total there are 1M patches each for Sentinel-1 GRD, Sentinel-2 L1C and Sentinel-2 L2A, resulting in 1.5TB as three tar.gz files.

\textbf{Does the dataset contain all possible instances or is it a sample (not necessarily random) of instances from a larger set?} The dataset is a sample of all Sentinel-1/2 satellite images. While the dataset can still be extended, we make it as representative as possible by ensuring global coverage, multiple modalities and multiple timestamps.

\textbf{What data does each instance consist of?} Sentinel-1/2 images along with meta-data captured from the space.

\textbf{Is there a label or target associated with each instance?} No, our dataset is unlabeled. However, each patch is bound with geographical location and acquisition time, thus a match to other labeled maps is possible.

\textbf{Is any information missing from individual instances?} No.

\textbf{Are relationships between individual instances made explicit (e.g., users’ movie ratings, social
network links)?} Not applicable, though geographic location / acquisition time / product type / other properties can be extracted if needed.

\textbf{Are there recommended data splits (e.g., training, development/validation, testing)?} The dataset is intended for unsupervised pre-training. Users are free to use either the full split or a subset (either a subset of modalities or a subset of geographical patches) based on their targeted applications.

\textbf{Are there any errors, sources of noise, or redundancies in the dataset?} Yes, as mentioned in the data collection section, there are two kinds of noise/redundancies: first, potential overlap around grid cell boundaries; second, potential noise of clouds from inaccurate cloud filtering.

\textbf{Is the dataset self-contained, or does it link to or otherwise rely on external resources (e.g.,
websites, tweets, other datasets)?} The dataset is self-contained.

\textbf{Does the dataset contain data that might be considered confidential (e.g., data that is protected
by legal privilege or by doctor-patient confidentiality, data that includes the content of individuals’ non-public communications)?} No.

\textbf{Does the dataset contain data that, if viewed directly, might be offensive, insulting, threatening,
or might otherwise cause anxiety?} No.

\textbf{Does the dataset identify any subpopulations (e.g., by age, gender)?} No.

\textbf{Is it possible to identify individuals (i.e., one or more natural persons), either directly or indirectly (i.e., in combination with other data) from the dataset?} No.

\textbf{Does the dataset contain data that might be considered sensitive in any way (e.g., data that reveals race or ethnic origins, sexual orientations, religious beliefs, political opinions or union memberships, or
locations; financial or health data; biometric or genetic data; forms of government identification, such as social security numbers; criminal history)?} No.

\subsection{Collection process}

\textbf{How was the data associated with each instance acquired?} The data was collected from the publicly available Sentinel-1/2 database. 

\textbf{What mechanisms or procedures were used to collect the data (e.g., hardware apparatus or
sensor, manual human curation, software program, software API)?} Google Earth Engine with Python were used to collect the data.

\textbf{If the dataset is a sample from a larger set, what was the sampling strategy (e.g., deterministic,
probabilistic with specific sampling probabilities)?} The patch locations are Gaussian sampled around a city center (50km) which is uniformly sampled from top-10k populated cities across the globe. The timestamps are sampled from four seasons (dates around Mar 20th, Jun 21st, Sep 22nd and Dec 21st) in the year 2020/2021.

\textbf{Who was involved in the data collection process (e.g., students, crowdworkers, contractors) and how were they compensated (e.g., how much were crowdworkers paid)?} The data was automatically collected and verified by the authors.

\textbf{Over what timeframe was the data collected?} The data was collected by the authors between February and March 2022. The images within the dataset were captured in the year 2020/2021.

\textbf{Were any ethical review processes conducted (e.g., by an institutional review board)?} No.

\textbf{Did you collect the data from the individuals in question directly, or obtain it via third parties or other sources (e.g., websites)?} The data was collected from open sources.

\textbf{Were the individuals in question notified about the data collection?} N/A.

\textbf{Did the individuals in question consent to the collection and use of their data?} N/A.

\textbf{If consent was obtained, were the consenting individuals provided with a mechanism to revoke their consent in the future or for certain uses?} N/A.

\textbf{Has an analysis of the potential impact of the dataset and its use on data subjects (e.g., a data protection impact analysis) been conducted?} N/A.

\subsection{Preprocessing/cleaning/labeling} 

\textbf{Was any preprocessing/cleaning/labeling of the data done (e.g., discretization or bucketing, tokenization, part-of-speech tagging, SIFT feature extraction, removal of instances, processing of missing values)?} The data was pre-processed online during the collection/downloading process: filtering out cloudy patches and overlapping patches. No further pre-processing was done.

\textbf{Was the “raw” data saved in addition to the preprocessed/cleaned/labeled data (e.g., to support unanticipated future uses)?} No. The cloudy and overlapping patches were removed before downloading.

\textbf{Is the software used to preprocess/clean/label the instances available?} Yes, we use Google Earth Engine with Python which is freely available.

\subsection{Uses}

\textbf{Has the dataset been used for any tasks already?} In this paper we use the dataset to benchmark several self-supervised methods on several downstream tasks. 

\textbf{Is there a repository that links to any or all papers or systems that use the dataset?} Yes we will organize and maintain all related information at \url{https://github.com/zhu-xlab/SSL4EO-S12}.

\textbf{What (other) tasks could the dataset be used for?} The main function of this dataset is to provide a pre-training dataset for both the study of self-supervised learning, and specific downstream applications. The dataset can also be used as a baseline for further pre-training datasets in Earth observation. In addition, the dataset can be used directly for applications like image retrieval, domain adaptation and style transfer.

\textbf{Is there anything about the composition of the dataset or the way it was collected and preprocessed/cleaned/labeled that might impact future uses?} We do not unify the orbiting (ascending/descending) of Sentinel-1 data, which should be taken into consideration for SAR related applications. However, the orbiting information can be found in the meta-data and the dataset can be further processed for targeting applications.

\textbf{Are there tasks for which the dataset should not be used?} The authors are not aware of any specific task that should be avoided.

\subsection{Distribution}

\textbf{Will the dataset be distributed to third parties outside of the entity (e.g., company, institution,
organization) on behalf of which the dataset was created?} Yes, the dataset is publicly available.

\textbf{How will the dataset will be distributed (e.g., tarball on website, API, GitHub)?} The dataset is distributed as tarball on mediaTUM. Access to the dataset can be found at \url{https://github.com/zhu-xlab/SSL4EO-S12}.

\textbf{When will the dataset be distributed?} Starting from June 2022.

\textbf{Will the dataset be distributed under a copyright or other intellectual property (IP) license, and/or under applicable terms of use (ToU)? } CC-BY.

\textbf{Have any third parties imposed IP-based or other restrictions on the data associated with the instances?} No.

\textbf{Do any export controls or other regulatory restrictions apply to the dataset or to individual
instances?} No.

\subsection{Maintenance} 

\textbf{Who is supporting/hosting/maintaining the dataset?} The dataset is hosted by mediaTUM and supported/maintained by the authors.

\textbf{How can the owner/curator/manager of the dataset be contacted (e.g., email address)?} The authors can be reached at their email addresses: \{yi.wang, nassim.aitalibraham, conrad.albrecht, chenying.liu\}@dlr.de, and \{zhitong.xiong, xiaoxiang.zhu\}@tum.de.

\textbf{Is there an erratum?} If errors are found an erratum will be added.

\textbf{Will the dataset be updated (e.g., to correct labeling errors, add new instances, delete instances)?} Any updates will be posted and the dataset will be versioned.

\textbf{If the dataset relates to people, are there applicable limits on the retention of the data associated with the instances (e.g., were individuals in question told that their data would be retained for a fixed period of time and then deleted)?} N/A.

\textbf{Will older versions of the dataset continue to be supported/hosted/maintained?} Depending on the updates (if there are), we will either continue hosting the older versions or make a clear update log that older versions can be generated from the newest version.

\textbf{If others want to extend/augment/build on/contribute to the dataset, is there a mechanism for
them to do so?} Yes, please feel free to reach out to us.

\subsection{Author statement of responsibility}

The authors confirm all responsibility in case of violation of rights and confirm the licence associated with the dataset.

\clearpage
{
\printbibliography[segment=1]
}

\end{document}